%% file: CVPR_2025/main.tex
\documentclass[10pt,twocolumn,letterpaper]{article}

\usepackage[9]{cvpr} 

\usepackage{xcolor,colortbl}
\usepackage{pifont}         

\usepackage{graphicx} 
\usepackage{xspace}
\usepackage{multirow}
\usepackage{multicol}
\usepackage{colortbl} 
\usepackage[utf8]{inputenc} 
\usepackage[T1]{fontenc}    
\usepackage{url}            
\usepackage{booktabs}       
\usepackage{amssymb}
\usepackage{amsmath}%
\usepackage{amsfonts}%
\usepackage{amsthm}
\usepackage{microtype}      
\usepackage{bbm}
\usepackage{amsmath}
\usepackage{algpseudocode}
\usepackage[ruled,vlined]{algorithm2e}
\usepackage{wrapfig}

\definecolor{myblue}{RGB}{59,103,188}
\definecolor{mygray}{RGB}{120,120,120}
\definecolor{myyellow}{RGB}{255,184,0}

\newcommand{\tteq}{\text{\,\texttt{=}\,}}

\theoremstyle{definition}

\newtheorem{defn}{Definition}

\newtheorem{remark}{Remark}

\newtheorem{prop}{Proposition}

\definecolor{cvprblue}{rgb}{0.21,0.49,0.74}
\usepackage[pagebackref,breaklinks,colorlinks,allcolors=cvprblue]{hyperref}

\def\paperID{17057} 
\def\confName{CVPR}
\def\confYear{2025}

\title{Multi-Group Proportional Representation for Text-to-Image Models}

\author{Sangwon Jung$^{1}$ ~~ Alex Oesterling$^{2}$ ~~ Claudio Mayrink Verdun$^2$ \\ 
Sajani Vithana$^{2}$ ~ Taesup Moon$^{1,3}$ ~ Flavio P. Calmon$^{2}$\vspace{.1in} \\
$^1$ ECE, Seoul National University \\ 
$^2$ Harvard University \ \ \ \ 
$^3$ ASRI/INMC/IPAI/AIIS, Seoul National University
}

\begin{document}
\maketitle
\input{CVPR_2025/sec/0_abstract}    
\input{CVPR_2025/sec/1_intro}
\input{CVPR_2025/sec/2_relatedworks}

\input{CVPR_2025/sec/3_mpr_def}

\input{CVPR_2025/sec/4_experiments}
\input{CVPR_2025/sec/5_method}

\input{CVPR_2025/sec/9_conclusion}

{
    \small
    \bibliographystyle{ieeenat_fullname}
    \bibliography{bib}
}

\input{CVPR_2025/sec/X_suppl}


\end{document}

%% file: CVPR_2025/sec/0_abstract.tex
\begin{abstract}
Text-to-image (T2I) generative models can create vivid, realistic images from textual descriptions. As these models proliferate, they expose new concerns about their ability to represent diverse demographic groups, propagate stereotypes, and efface minority populations. Despite growing attention to the ``safe'' and ``responsible'' design of artificial intelligence (AI), there is no established methodology to systematically measure and control representational harms in image generation. This paper introduces a novel framework to measure the representation of intersectional groups in images generated by T2I models by applying the Multi-Group Proportional Representation (MPR) metric. 
MPR evaluates the worst-case deviation of representation statistics across given population groups in images produced by a generative model, allowing for flexible and context-specific measurements based on user requirements. We also develop an algorithm to optimize T2I models for this metric. 
Through experiments, we demonstrate that MPR can effectively measure representation statistics across multiple intersectional groups and, when used as a training objective, can guide models toward a more balanced generation across demographic groups while maintaining generation quality.  \footnote{The code is available at \href{https://github.com/sangwon-jung94/mpr-t2i}{https://github.com/sangwon-jung94/mpr-t2i}}

\end{abstract}

%% file: CVPR_2025/sec/1_intro.tex
\section{Introduction}
\label{sec:intro}

Recent text-to-image (T2I) generative models, such as Google Gemini \cite{google_gemini}, DALL-E \cite{ramesh2021zero}, Midjourney \cite{midjourney}, Adobe Firefly \cite{adobe_firefly} and Stable Diffusion \cite{rombach2022high}, have pushed the boundaries of creativity with AI, offering new possibilities in fields ranging from digital art to product design. 
However, as these technologies become more prevalent, there has been a surge in concerns about their portrayal of diverse population groups \cite{bianchi2023easily}, their potential biases \cite{cho2023dall}, and their ability to promote stereotypes \cite{mack2024they,fraser2023diversity}.
A notable example is the controversy surrounding Google's Gemini AI, which produced historically inaccurate images \cite{geminiarticle} and received significant attention in the press \cite{geminiarticle2}. 

\begin{figure*}[t!]
    \centering    
    \includegraphics[width=0.8\textwidth]{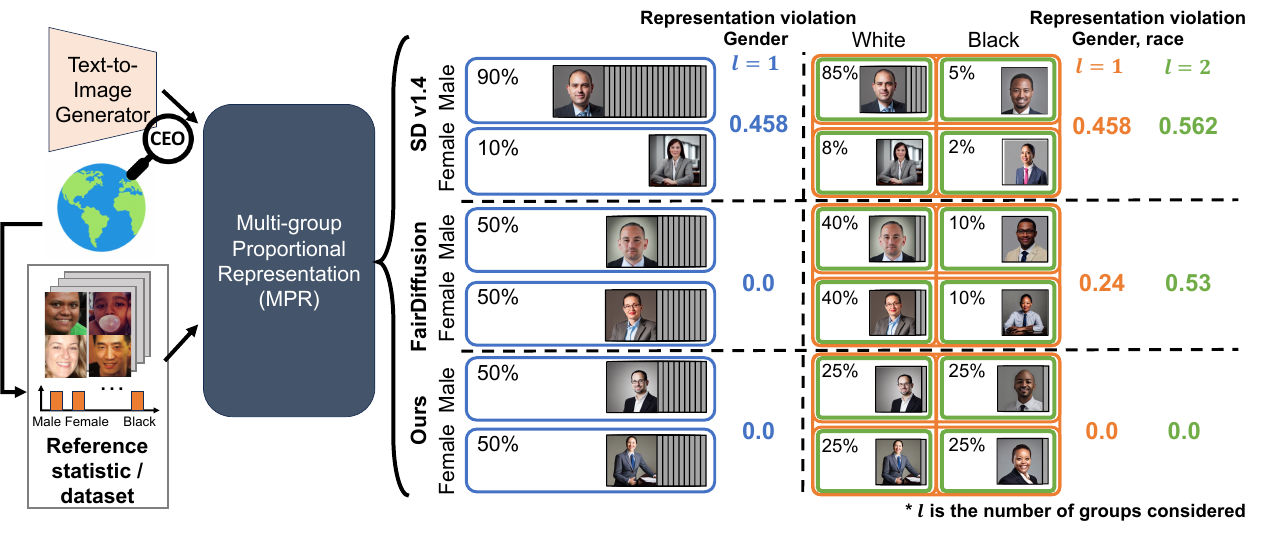}
    \caption{\small Overview of the Multi-Group Proportional Representation (MPR) framework for evaluating text-to-image generators. Given a prompt (e.g., ``CEO''), the framework compares the distribution of demographic attributes (such as gender and race) between generated images and a reference dataset or target statistics. MPR measures the maximum deviation in representation across multiple intersectional groups, enabling systematic quantification of representational biases in generated images.}
    \label{fig:teaser}
\end{figure*}

The challenge of achieving fair representation in generative AI requires, first and foremost, a precise methodology for \emph{measuring} representation. However, current approaches that assess representation in T2I models face several limitations surrounding how they define and measure \emph{representational fairness} \cite{wan2024survey}. Most studies \cite{friedrich2023fair, cho2023dall, wang2024concept} focus on a single axis of outward-facing identity representation, such as gender, overlooking the complex interplay of multiple identity factors. While some research attempts to consider multiple axes of representation, e.g., \cite{bianchi2023easily,shen2023finetuning, zhang2023iti, yesiltepe2024mist}, there is a lack of systematic measurement methods, particularly for intersectional groups. Determining appropriate representation goals is also challenging. Recent frameworks like BIGbench \cite{luo2024bigbench} and OpenBias \cite{d2024openbias} have proposed more systematic approaches to evaluate bias in text-to-image models by leveraging multi-modal LLMs. However, these methods still lack a theoretically grounded framework for measuring representational fairness across intersectional groups and optimizing generative models accordingly.

Our proposed framework addresses these limitations by providing a unified mathematical framework that can measure and mitigate representational harm. For instance, sometimes we may want equal representation (such as generating images of a CEO with equal gender representation), and in others, we may care about reflecting some real-world or historically accurate distribution of identities (such as not generating racially diverse 1943 German soldiers as done by Gemini \cite{geminiarticle}).
\emph{In this work, we apply a metric to T2I models that can address both of these needs and allow us to improve representation in image generation.} 
Our approach represents the comprehensive framework that can flexibly evaluate representation across various complexities of intersectional groups while being adaptable enough to accommodate both equal and proportional representation goals, providing a principled foundation for measuring and mitigating representational harm in T2I systems. Indeed, we empirically show that optimizing a generative image model with MPR improves the representation of its generations (Fig. \ref{fig:teaser}). 


This paper contributes to the field of fairness in generative AI by introducing the use of Multi-group Proportional Representation (MPR), initially proposed for retrieval tasks \cite{oesterling2024multi}, as a general metric for measuring and improving representation in T2I models. This extension provides a robust and flexible approach to quantifying fair proportional representation with respect to general computationally-identifiable groups. Importantly, our proposed MPR metric can be easily adjusted to accommodate complex or newly emerging definitions of group identity, making it a versatile tool for measuring representational harm in T2I models that can navigate the aforementioned tensions between factuality and representation. Our main contributions are as follows:
\begin{itemize}
    \item We introduce the Multi-Group Proportional Representation (MPR) metric, providing the first theoretically grounded framework for measuring and optimizing for representational fairness in T2I models. While MPR was originally proposed for retrieval tasks, we show how it can effectively capture intersectional representation in generated content with remarkable flexibility in handling different demographic attributes, group definitions, and fairness criteria.
    \item We demonstrate the scalability and effectiveness of MPR through experiments with state-of-the-art T2I models, showcasing its ability to capture biases related to various facets of identity. In particular, we use MPR to identify critical representation gaps for popular models in professional representation, trait attribution, and disability portrayal. We additionally show that MPR scales to a large set of attributes and adapts to query-contextual representation evaluation.
    \item We demonstrate how MPR can be used as a training objective to optimize T2I models for fair representation. Our experiments show that MPR-based fine-tuning achieves a more balanced demographic representation than existing debiasing methods while maintaining image quality and semantic accuracy.
\end{itemize}


The paper is organized as follows. In Section \ref{sec:related_works}, we present a comprehensive analysis of related work. In Section \ref{sec:MPR_metric}, we introduce MPR and discuss its theoretical properties in a generative context, including its ability to capture nuanced representational harm of intersectional groups. In Section \ref{sec:experiments}, through extensive experiments, we demonstrate the practical effectiveness of MPR in capturing representational harm across various demographic groups and their intersectional identities. In Section \ref{sec:intervention_methods}, we illustrate how MPR can be used to fine-tune a text-to-image generation model to mitigate the harm. 

%% file: CVPR_2025/sec/2_relatedworks.tex
\section{Related Work}
\label{sec:related_works}

\paragraph{Representational harm in T2I systems.} Recent research has highlighted key areas where T2I generation systems exhibit representational harm across various societal and demographic factors. One prominent area of concern is the representation of professions. Multiple studies have identified significant issues in how these systems depict various occupations, particularly related to gender \cite{friedrich2023fair, orgad2023editing, kim2023stereotyping}, race \cite{luccioni2024stable}, skin-tone \cite{cho2023dall, zhang2023iti} or age \cite{shen2023finetuning, zhang2023iti}. Some studies have shown that these issues extend to traits described by adjectives (\eg, ``poor'') and certain nouns (\eg, ``thug'') \cite{bianchi2023easily, fraser23friendly}. Research has also identified unfair representation in disability \cite{mack2024they} -- including emotional expression and representation of objects like wheelchairs -- and socioeconomic representation, where certain economic conditions are disproportionately associated with specific groups \cite{fraser2023diversity, bianchi2023easily}. With an increasing understanding of the wide range of unfairness potentially present in T2I generation systems, \cite{chinchure2024tibet} proposes a method to automate unfairness identification and dynamically generates potential areas of unfairness risk based on the prompt used to generate an image.

\textbf{Fairness metrics and evaluation methods.} Prior bias evaluation approaches in T2I models have various limitations, such as using qualitative evaluations with humans to assess the presence and impact of biases \cite{bansal2022well}. When examining a single type of bias, researchers often use a uniform distribution (e.g., each group represented equally) as a reference point for comparison \cite{cho2023dall, bansal2022well}. This approach provides a baseline against which deviations can be measured, helping to identify specific biases within the model outputs. However, the complexity of bias necessitates the consideration of multiple aspects simultaneously, as demonstrated in several studies \cite{shen2023finetuning, bianchi2023easily, luccioni2024stable, zhang2023iti, ovalle2023factoring}. Notably, quantitative metrics for assessing bias in intersectional groups remain limited, with only a few metrics currently available \cite{zhang2023iti, dinca2024openbias, shen2023finetuning,luo2024bigbench}. However, these metrics lack a systematic framework for determining the level of granularity of intersectionality in measuring representation and for selecting appropriate representation targets.


\textbf{Bias mitigation strategies.} Recent research has explored various approaches to mitigate bias in T2I models. Ethical statement prompting \cite{bansal2022well} employs post-processing and manual prompt tuning by appending statements like ``if all individuals can be a lawyer irrespective of their gender'' to prompts. \cite{friedrich2023fair} introduced FairDiffusion, a post-processing method that modifies text guidance using a lookup table to add text guidance vectors of minor groups while subtracting those of major groups. DebiasVL \cite{chuang2023debiasing}, TIME \cite{orgad2023editing} and UCE \cite{gandikota2024unified} utilize linear projections to filter out identity-related information from text embedding vectors. \cite{li2023fair} developed FairMapping, another post-processing method using linear projection of text embedding vectors. \cite{kim2023stereotyping} explored soft prompt tuning as an efficient training method for debiasing. More recently, some works proposed more sophisticated methods for debiasing by adjusting features fed into attention modules \cite{parihar2024balancing, yesiltepe2024mist}, utilizing reinforcement learning \cite{miao2024training} or optimal transport-based regularization \cite{shen2023finetuning}. \cite{li2024controlling} tilted the distribution of training and generated data through a pseudo-density to adjust the fidelity and diversity of generative models. While not specifically focused on T2I systems, \cite{kim2024training} introduced DSM, an in-processing re-weighting learning scheme optimized for diffusion models, and \cite{choi2024fair} developed FairSwitch, a post-processing method that modifies samples in the middle of the backward process of diffusion models. Finally, \cite{zhao2022scaling} propose a method in predictive settings to scale fair learning to large numbers of intersectional groups using knowledge distillation.

%% file: CVPR_2025/sec/3_mpr_def.tex
\section{Multi-group Proportional Representation in Image Generation}\label{sec:MPR_metric}

Our goal is to generate images that are accurate with respect to a given prompt while being representative across various intersectional population groups in a given social or historical context. 
As mentioned in the Introduction, we draw upon the MPR metric, first introduced in retrieval systems \cite{oesterling2024multi}, and extend it to a T2I generation.  The formal definition is provided in Definition~\ref{def1}.


\begin{defn}\label{def1}
    Let $X_g\in\mathbb{R}^d$ for some $d<\infty$ be the random variable representing a generated image from a text-to-image generative model, with a distribution $G_q$, conditioned on a given prompt $q$. Let $X_r\in\mathbb{R}^d$ be the random variable representing an image from a reference population with a distribution $R$. For a bounded class of functions $\mathcal{C}$, where  $c:\mathbb{R}^d\rightarrow\mathbb{R}$ for each $c\in\mathcal{C}$, the MPR metric is defined as:
    \begin{align}\label{mpr_def}
        \text{MPR}(\mathcal{C},G_q,R) \triangleq \sup_{c\in\mathcal{C}} \big|\mathbb{E}_{G_q} [c(X_g)] - \mathbb{E}_{R}[c(X_r)]\big|.
    \end{align}
\end{defn}
For $X_g$ and $X_r$ in \eqref{mpr_def}, one can use the direct pixel values, $d$-dimensional vector embeddings of the respective images, or attribute vectors indicating the sample's group membership with respect to the sensitive attributes being considered. In this work, we opt for the attribute vectors to conduct experiments with reduced noise in the data.


We note that MPR has appeared in different contexts under different names, e.g., it is also called the maximum mean discrepancy (MMD) \cite{mmd1,mmd2} of a selected set of representation statistics determined by $\mathcal{C}$, with respect to the generative and reference distributions. It can be seen as a special case of integral probability metrics (IPM) \cite{peyre2019computational,ipm1,ipm2}. In fairness literature, MPR builds off of work in multicalibration \cite{hebert2018multicalibration} and multiaccuracy \cite{kim2019multiaccuracy} for multi-group fair classification. 


The MPR metric is a generalization of many of the representational fairness metrics for image generation used in prior studies. For instance, by selecting 1) $X_g, X_r\in\{0,1\}$ as the binary membership in a single group type, such as gender or race, 2) uniformly distributed $R$ over $\{0,1\}$, and 3) $\mathcal{C}=\{c_0,c_1\}$, where $c_0(X)=\mathbf{1}_{\{X=0\}}$ and $c_1(X)=\mathbf{1}_{\{X=1\}}$ with $\mathbf{1}(\cdot)$ denoting the identity function, MPR computes the maximum difference between the probabilities across groups, as done in \cite{friedrich2023fair, gandikota2024unified}. Similarly, when $X_g, X_r$ are vectors representing membership in multiple groups, and $\mathcal{C}$ corresponds to a set of full-depth decision trees, MPR measures unfairness across all possible intersectional groups as in \cite{shen2023finetuning, zhang2023iti}.

The key advantage of MPR over the existing metrics lies in its flexibility regarding the choice of reference distribution and function class $\mathcal{C}$. By employing context-aware reference distributions that adapt to the prompt, MPR can address the \emph{diversity versus prompt relevance} tradeoff encountered previously by Gemini \cite{geminiarticle}. For example, if the prompt is \emph{“an image of a 2022 FIFA World Cup athlete”}, a representation metric should not penalize the system for failing to generate images of female athletes, as this aligns with the real-world context of events that took place in 2022. To achieve this, we can measure MPR against a reference distribution $R_q$ conditioned on the prompt, ensuring that the evaluation focuses only on a contextual and relevant reference distribution.

Additionally, MPR allows control over the granularity of representational difference measurements by adjusting the complexity of the function class. Some users may prefer to measure deviations between generated and reference distributions with high precision, using complex function classes like full-depth decision trees or deep neural networks. However, as the number of group attributes grows or when continuous embedding vectors are involved, it may become necessary to reduce the complexity.
For instance, as in \cite{monteiro2022epistemic}, considering 27 group attributes results in over 7 billion intersectional groups, requiring a number of samples on the order of the human population to measure representation effectively. This makes it impractical to accurately measure the degree of unfairness at the finest granularity of intersectionality. In such cases, MPR can navigate the intersectionality-complexity tradeoff by using simpler function classes, like linear classifiers or low-depth decision trees. A more detailed mathematical discussion on the impact of function complexity on MPR is provided in Section \ref{sec:MPR_metric}.

\begin{remark} A main challenge in the MPR framework is determining the reference distribution $R$. Ideally, $R$ should be well-balanced across intersectional groups, representing a diverse population. Since such complex distributions rarely have been known in straightforward analytic forms, we approximate $R$ by randomly sampling from a curated dataset $\mathcal{D}$ that has ideal representation statistics. An example of $\mathcal{D}$ is the FairFace dataset \cite{karkkainen2021fairface}, which is specifically designed to be balanced across race, gender, and age groups. We could also imagine sampling individuals according to the Census data of a given country, such as US Bureau of Labor Statistics \citep{us_bureau} data, as well as demographic information from other relevant international institutes and entities, such as the United Nations Population Division \cite{UN_populationdivision}, the World Bank \cite{worldbank_data}, Eurostat \cite{eurostats_data}, ILOSTAT \cite{ilostat_data}, or the Organisation for Economic Co-operation and Development (OECD) \cite{oecd_data}. In Section~\ref{sec:experiments}, we discuss how we select $\mathcal{D}$ in our experiments. 
\end{remark}


\subsection{MPR with Empirical Distributions}
We approximate MPR in \eqref{mpr_def} with empirical distributions, by generating $k$ images from $G_q$, and sampling $m$ images from a reference dataset with distribution $R$:
\begin{align}\label{emp_mpr_def}
        \text{MPR}(\mathcal{C},\widehat{G}_q,\widehat{R}) &= \sup_{c\in\mathcal{C}} \left|\frac{1}{k}\sum_{i=1}^k c(x_i^g) - \frac{1}{m}\sum_{j=1}^m c(x_j^r)\right|
\end{align}
where $\widehat{G}_q=\{x_i^g\}_{i = 1}^{k}$ and $\widehat{R}=\{x_j^r\}_{j = 1}^{m}$ are the generated and reference samples used to approximate $G_q$ and $R$. To quantify the closeness between the true MPR defined in \eqref{mpr_def} and its empirical approximation in \eqref{emp_mpr_def}, we derive bounds for the approximation error in Proposition~\ref{gen}.

\begin{prop}[Generalization Gap of MPR]\label{gen}  
For any given prompt $q$, let $\widehat{G}_q$ and $\widehat{R}$ be the sets of generated and reference samples used to approximate $G_q$ and $R$, respectively.
Then, for any bounded class of functions $\mathcal{C}$ and any $\delta>0$,
\begin{align}
    &\left|\mathrm{MPR}(\mathcal{C},\widehat{G}_q,\widehat{R})-\mathrm{MPR}(\mathcal{C},G_q,R)\right| \nonumber\\
    &\qquad\qquad\leq 2\mathcal{R}_{\widehat{G}_q}(\mathcal{C}) + 2\mathcal{R}_{\widehat{R}}(\mathcal{C}) + B\sqrt{\frac{\log\left( \frac{2}{\delta}\right)}{2(k+m)}} \label{eq:prop1}
\end{align}
with probability at least $1 - \delta$, where $B=\sup_{\substack{c\in\mathcal{C}\\X\neq X'}}|c(X)-c(X')|$ and $\mathcal{R}_\mathcal{A}(\mathcal{C})$ is the Rademacher complexity of $\mathcal{C}$ relative to the set $\mathcal{A}$.
\end{prop}

Proposition \ref{gen} reveals the fundamental trade-off among three key factors: the approximation error, the sample sizes of $\widehat{G}_q$ and $\widehat{R}$, and the complexity of function class $\mathcal{C}$. The approximation error decreases as we reduce the complexity of $\mathcal{C}$ and increase the sample sizes of $\widehat{G}_q$ and $\widehat{R}$. While Proposition~\ref{gen} offers guidance for selecting appropriate sample sizes to achieve a target error threshold with a fixed $\mathcal{C}$, we extend our analysis in two directions. First, we present empirical results in Section \ref{appen:maximum_gap} that demonstrate how the approximation error varies with the complexity of $\mathcal{C}$. Second, we discuss practical heuristics for optimal selection of $k$ and $m$ in Section \ref{appen:practical_guidelines}. Beyond representational harm related to gender, age, and race, there are also unfairness tied to factors like disabilities, emotions, and image-backgrounds, as qualitatively discussed by \cite{bianchi2023easily, mack2024they}. Increasing the complexity of function class $\mathcal{C}$ enables MPR to capture and measure these subtle representational disparities with greater precision. Guided by Proposition~\ref{gen}, we can methodically determine the sample sizes needed to achieve desired measurement accuracy for a given complexity of $\mathcal{C}$. This approach allows MPR to reliably quantify both conventional demographic and more nuanced representational disparities.

Thus far, we have considered MPR for a single prompt-conditioned generative distribution, $G_q$. Next, we approximate the average MPR of the T2I generation model over the entire prompt space by leveraging an empirical distribution of sampled prompts. Proposition \ref{gen_prompt} provides a bound on the difference between the empirical mean of MPR across $N$ prompts and the true expected MPR, ensuring that MPR generalizes to new, unseen prompts.


\begin{prop}[Generalization of MPR over all Prompts]\label{gen_prompt} Let $P$ denote the distribution of prompts, and $\{q_1,\dotsc,q_N\}$ denote a set of independent prompts sampled from $P$. Then, for all $\epsilon > 0$,
\begin{align}
    &\mathbb{P}\!\left(\left|\!\frac{1}{N}\sum_{i=1}^N \!\mathrm{MPR}(\mathcal{C},\!\widehat{G}_{q_i},\!\widehat{R})\!-\!\mathbb{E}_{Q\sim P}[\mathrm{MPR}(\mathcal{C},\!G_{Q},\!R)]\right|\!\!\geq\!\epsilon\!\right)\nonumber\\
    &\leq \exp\left(-\frac{\epsilon^2N}{8}\right)+\exp\left(-\frac{2(k+m)}{B^2}\left(\frac{\epsilon}{2}-2\lambda\right)\right)
\end{align}
where $k$ and $m$ are the numbers of generated and reference samples used in the calculation of $\mathrm{MPR}(\mathcal{C},\widehat{G}_{q_i},\widehat{R})$, $Q$ is the random variable representing a prompt, $Q\sim P$, and $\lambda=\sup_{Q\sim P}\mathcal{R}_{\widehat{G}_{Q}}(\mathcal{C})+\mathcal{R}_{\widehat{R}}(\mathcal{C})$.    
\end{prop}

\subsection{MPR computation according to \texorpdfstring{$\mathcal{C}$}{} }
In this paper, we consider two main classes of functions to describe sensitive intersectional groups: 1) bounded linear functions and 2) binary decision trees. The following results characterize MPR for each function class. 

\begin{prop}\label{linear}
    For the class of bounded linear functions, $\mathcal{C}=\{c:w^Tx\big|w\in\mathbb{R}^d, \|w\| \leq 1\}$, 
    \begin{align}
        \text{MPR}(\mathcal{C},\widehat{G}_q,\widehat{R}) = \frac{1}{||\tilde{a}^TX||}\tilde{a}^TXX^T\tilde{a},
    \end{align}
    in which, $\tilde{a}\in\mathbb{R}^{k+m}$ has $i$-th entry given by $\tilde{a}_i = \mathbbm{1}_{i \leq k}\frac{1}{k}-\mathbbm{1}_{i>m}\frac{1}{m}$ and $X\in \mathbb{R}^{(k+m)\times d}$ is a matrix with its first $k$ rows representing the generated images $\{x_i^g\}_{i=1}^k$ and the last $m$ rows representing the reference images sampled $\{x_i^r\}_{i=1}^m$.
\end{prop}
Next, we present the closed form of MPR with binary decision trees. In this case, we assume $x_i^g,x_i^r\in\{-1,+1\}^n$ to be feature vectors denoting the respective sample's group membership (binary) in $n$ sensitive attributes. For example, if $n=2$ with the corresponding sensitive attributes being race and gender, an image of a person belonging to race 1 and gender 1 is denoted by the vector $x=[+1,+1]$. 

\begin{prop}\label{dt}
    For the class of binary decision trees of depth $\ell\leq n$, where $\mathcal{C}=\{c:\{-1,+1\}^n\rightarrow\{-1,+1\}\}$,
    \begin{align}
        \text{MPR}(\mathcal{C},\widehat{G}_q,\widehat{R}) = \max_{I\subset\{1,2,\dots,n\},|I|=\ell} 2\text{TV}(\tilde{G}_q^I,\tilde{R}^I)
    \end{align}
    where $\tilde{G}_q^I$ and $\tilde{R}^I$ are the marginal distributions corresponding to $\widehat{G}_q$ and $\widehat{R}$ over the attributes in set $I$, and $\text{TV}(A,B)$ denotes the total variation between distributions $A$ and $B$.
\end{prop}

%% file: CVPR_2025/sec/4_experiments.tex

\section{Quantitative Assessment of Representational Harm Using MPR}\label{sec:experiments}
We investigate representational harms inherent in various T2I models using the MPR metric. Unlike previous studies, e.g., \cite{cho2023dall, friedrich2023fair, shen2023finetuning}, which primarily focused on single-group such as occupational stereotypes, we extend the analysis to multiple demographic groups and quantify representational harms related to traits and disabilities, which were qualitatively studied in the previous works \cite{bianchi2023easily, mack2024they}. 

\noindent \textbf{Experimental setup}. 
For each concept under investigation, we generate a set of images using prompts such as \emph{``a portrait photo of a \{concept\}''}. The MPR metric is then calculated to assess how biased a T2I model’s representation is relative to a reference dataset in terms of multiple groups. We generate 1,000 images for each experiment and use bootstrapping with 1,000 resamples with replacement, performed 100 times, to compute the average MPR and its standard deviation. Unless specified otherwise, we use the Fairface \cite{karkkainen2021fairface} test dataset as our reference dataset, which is carefully designed to be balanced for gender, age, and race. 
The group labels of images for gender, age, and race are obtained by linear classifiers probed on CLIP embedding space with Fairface\footnote{We note a body of work discussing the ethics of predicting sensitive attributes \citep{hamidi2018gender}, but do not presume to address this issue here.}. When using decision trees as $\mathcal{C}$, the outputs of these classifiers are encoded as one-hot vectors, i.e., binary vectors where all elements are 0 except for one element, which is 1. We utilize three open-source T2I models, including Stable Diffusion v1.4 \cite{rombach2022high}, Stable Diffusion v2.1 \cite{SD2.1}, and SDXL \cite{podellsdxl}. We report MPR values for several intervention methods: the baseline models without any intervention (denoted here as Vanilla), Entigen \cite{bansal2022well}, FairDiffusion \cite{friedrich2023fair}, ITI-GEN \cite{zhang2023iti}, 
UCE \cite{gandikota2024unified}.
Unless otherwise specified, all baselines are adjusted to mitigate representational harm based on the reference used by MPR. UCE, Entigen, and FairDiffusion are implemented to address the unfairness only for gender, while ITI-GEN considers intersectional groups regarding gender, age, and race. The other implementation details are in Section \ref{appen:implementation_detail}.

\begin{figure}[t!]
    \centering
    \includegraphics[width=0.8\linewidth]{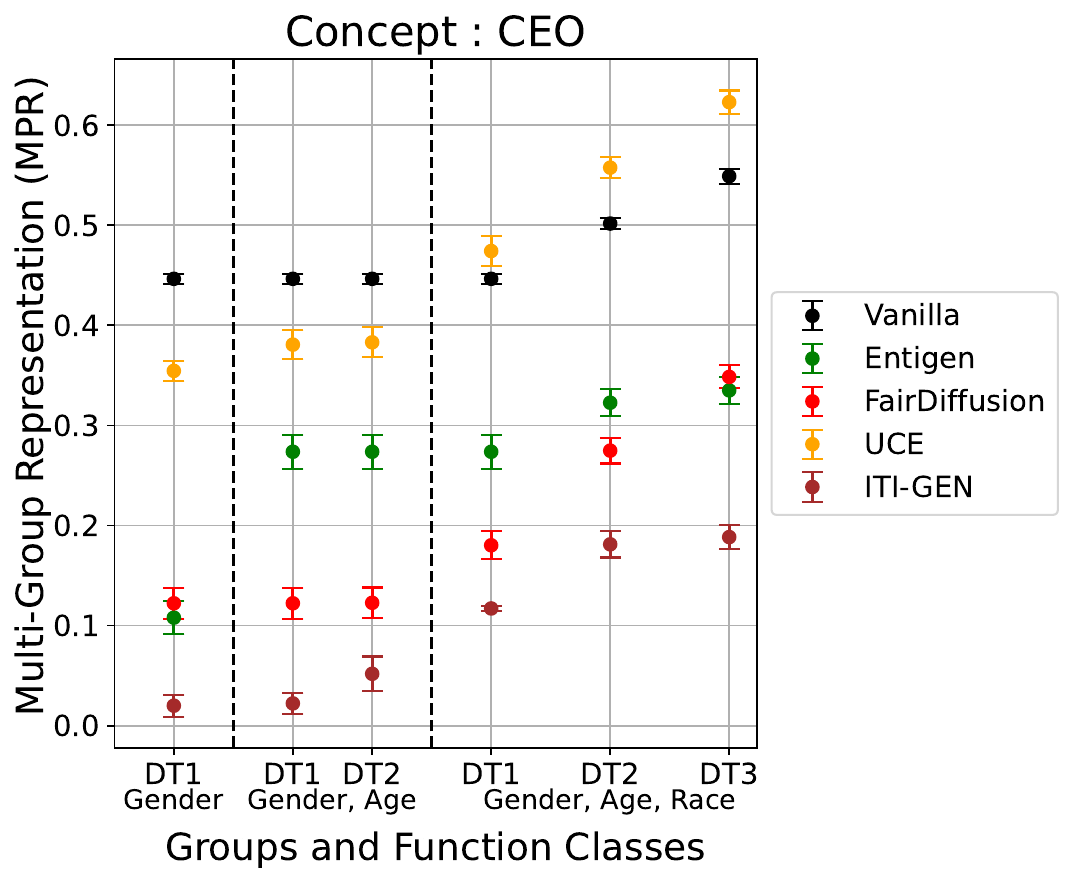}
    \caption{\small {\bf MPRs according to $C$ and group attributes when applying existing bsaeline methods to mitigate unfairness of SD v1.4 for ``CEO''.} DT represents decision trees, and the number following it indicates its depth. The text under DT refers to the group attributes used.}
    \label{fig:mpr_behavior}
\end{figure}

\input{CVPR_2025/tables/table1}

\subsection{Application of MPR across diverse scenarios} 

\paragraph{MPRs for profession representation.} 

Our MPR framework can adapt to a variety of scenarios, which differ in which groups are focused on and how fine-grained intersectional groups should be considered. To emulate these situations, we apply MPR while increasing the number of groups considered -- starting with gender and then adding age and race -- and simultaneously increasing the depth of the decision tree classifiers. 

Fig. \ref{fig:mpr_behavior} shows the MPR values for images generated by the vanilla SD v1.4 model and baseline methods for the prompts using ``CEO''. From the figure, we observe that as the number of group attributes and the depth of the decision trees (i.e., the complexity of the function class $\mathcal{C}$) increase, the MPR value increases. This is because MPR identifies the most over- or under-represented group from the reference population over  larger $C$. 
MPR exposes critical limitations in existing harm mitigation methods that were specifically developed to address gender disparity. While FairDiffusion, UCE, and Entigen demonstrate effective disparity reduction in their intended domain of gender fairness, their performance deteriorates significantly when considering intersectional attributes of age and race simultaneously. When considering race together, UCE's performance even falls below that of the vanilla model, highlighting the challenges of addressing multiple dimensions of representational harm beyond their original gender-focused design. These quantitative findings are visually corroborated by generated examples shown in Fig. \ref{fig:scratch_example} and \ref{fig:fairdiffusion_example}. For a comprehensive analysis across a wider range of occupations, we present additional MPR results in Section \ref{sec:intervention_methods}.
\vspace{-.5cm}

\paragraph{MPRs for trait representation.} MPR can calculate representational harm for any query, given the group estimation method for images. Specifically, we investigate the unfairness in intersectional groups for six human-related traits --``attractive'', ``emotional'', ``exotic'', ``poor'', ``terrorist'' and ``thug'' -- which were only qualitatively assessed in the previous research \cite{bianchi2023easily}\footnote{Here we used the same prompt of ``the portrait photo of a \{concept\}'' person as \cite{bianchi2023easily}.}. Table \ref{tab:mpr_trait} quantitatively presents the extent of intersectional group unfairness for these traits using MPR values across the three T2I models, considering gender, age, and race groups. We also report the features used as splits in the decision trees selected to compute MPR, showing that using decision trees provides interpretable results. It shows that all three models exhibit large MPR values for each trait, showing a significant degree of unfairness. Importantly, as shown in the results for the ``exotic'' trait, \emph{we can measure more severe unfairness when considering more fine-grained intersectional groups}. Finally, the splits used in the decision trees allow us to interpret which aspects of the T2I models exhibit the most unfairness. For example, in the cases of ``attractive'' and ``terrorist'', the splits mainly involve ``White'' and ``Middle Eastern'' groups, which align with the qualitative findings from \cite{bianchi2023easily}.

\input{CVPR_2025/tables/table2}
\vspace{-.5cm}
\paragraph{MPRs for disability representation.} The MPR framework's flexibility in accommodating different group estimators and reference datasets enables us to quantify diverse forms of representational harm, extending beyond traditional demographic categories like gender, age, and race to include more nuanced social dimensions. A compelling example of such subtle harm emerges in the context of disability representation, which Mack et al. \cite{mack2024they} explored in their analysis of images generated from the prompt \emph{"a photo of a person with a disability"}. Their work revealed a concerning pattern where generated images disproportionately depicted wheelchair users while underrepresenting individuals with visual, auditory, or cognitive disabilities. 
Building upon their qualitative observations, we provide a qualitative analysis using MPR. Our study focuses on the overrepresentation of mobility disabilities and its intersection with racial disparity. 
We use reference statistics from the U.S. Centers for Disease Control and Prevention (CDC) \cite{cdc}, which provide detailed data on the intersectional distribution of race and disability types. For tractability, we group  the disability into mobility disabilities (represented by wheelchair users) and non-mobility disabilities (encompassing cognitive, visual, and auditory impairments), and consider four major racial groups: White, Black, Asian, and Latino populations. To detect wheelchair presence in generated images, we employ the BLIP2 Visual Question Answering (VQA) system with the query \emph{``What objects are in the image?''}, deriving pseudo-labels from its responses. 
Table \ref{tab:mpr_disability} presents MPR values and their corresponding decision tree splits at depths 1 and 2. The analysis demonstrates that the three T2I models exhibit compounded bias: \emph{not only do they disproportionately generate images of wheelchair users over other forms of disability, but they also show substantial racial skew within these representations}. This is visually substantiated in Figure \ref{fig:disability_example}, where the SD v1.4 model frequently depicts white wheelchair users.

\subsection{Use cases unlocked by MPR}

The formulation of MPR and choice of $\mathcal{C}$ allows MPR to be used in settings where prior fairness metrics are unable to measure bias or fail to accurately capture the nuances of representation. We demonstrate a use case for querying a reference dataset to handle specific social and historical contexts and then discuss the benefits of exploring a general class of functions $\mathcal{C}$. 

\input{CVPR_2025/tables/table3}

\begin{figure}    
    \centering
    \begin{minipage}{0.45\linewidth}
        \centering
        \includegraphics[width=0.95\linewidth]{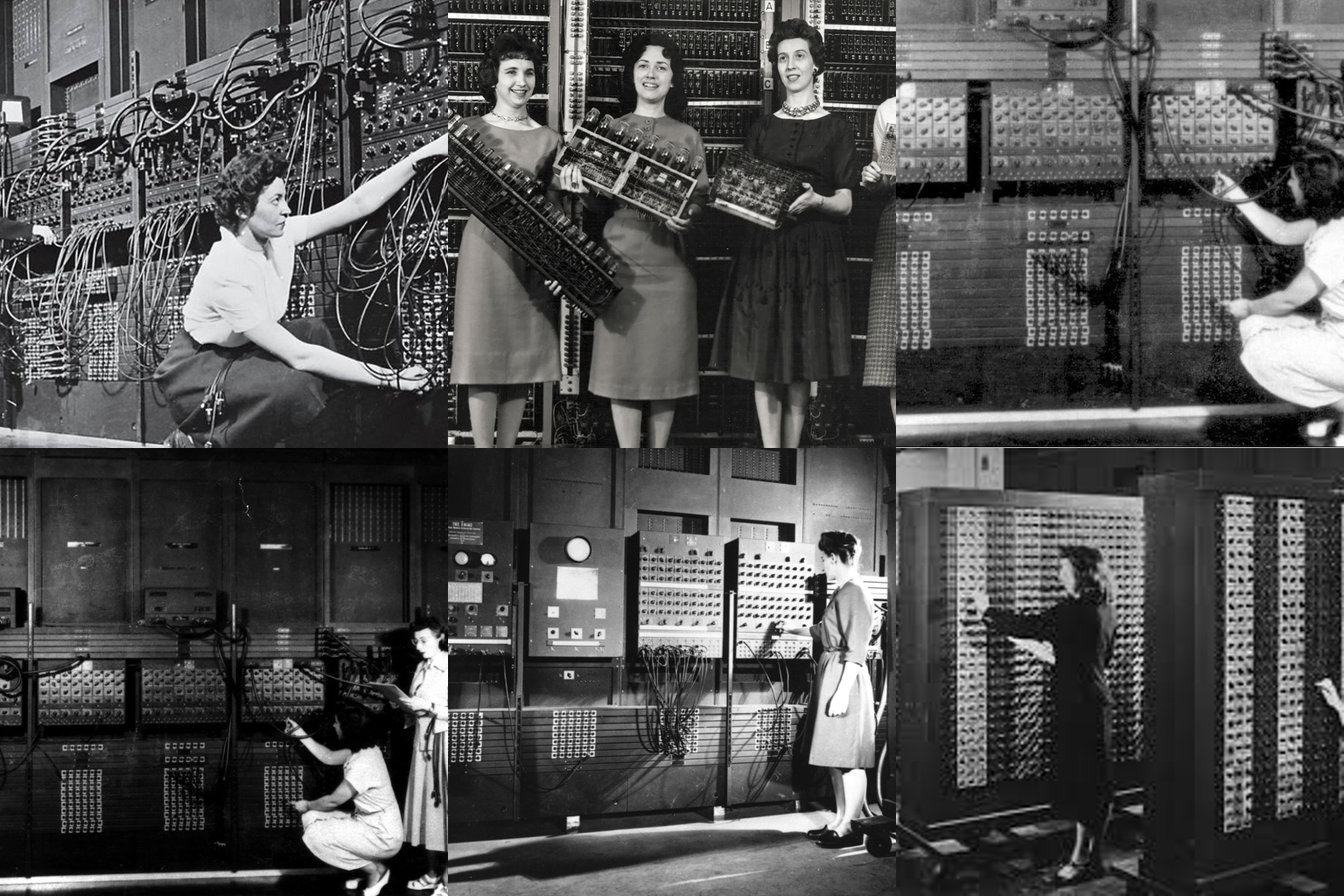}
    \end{minipage}%
    \begin{minipage}{0.45\linewidth}
        \centering
        \includegraphics[width=0.95\linewidth]{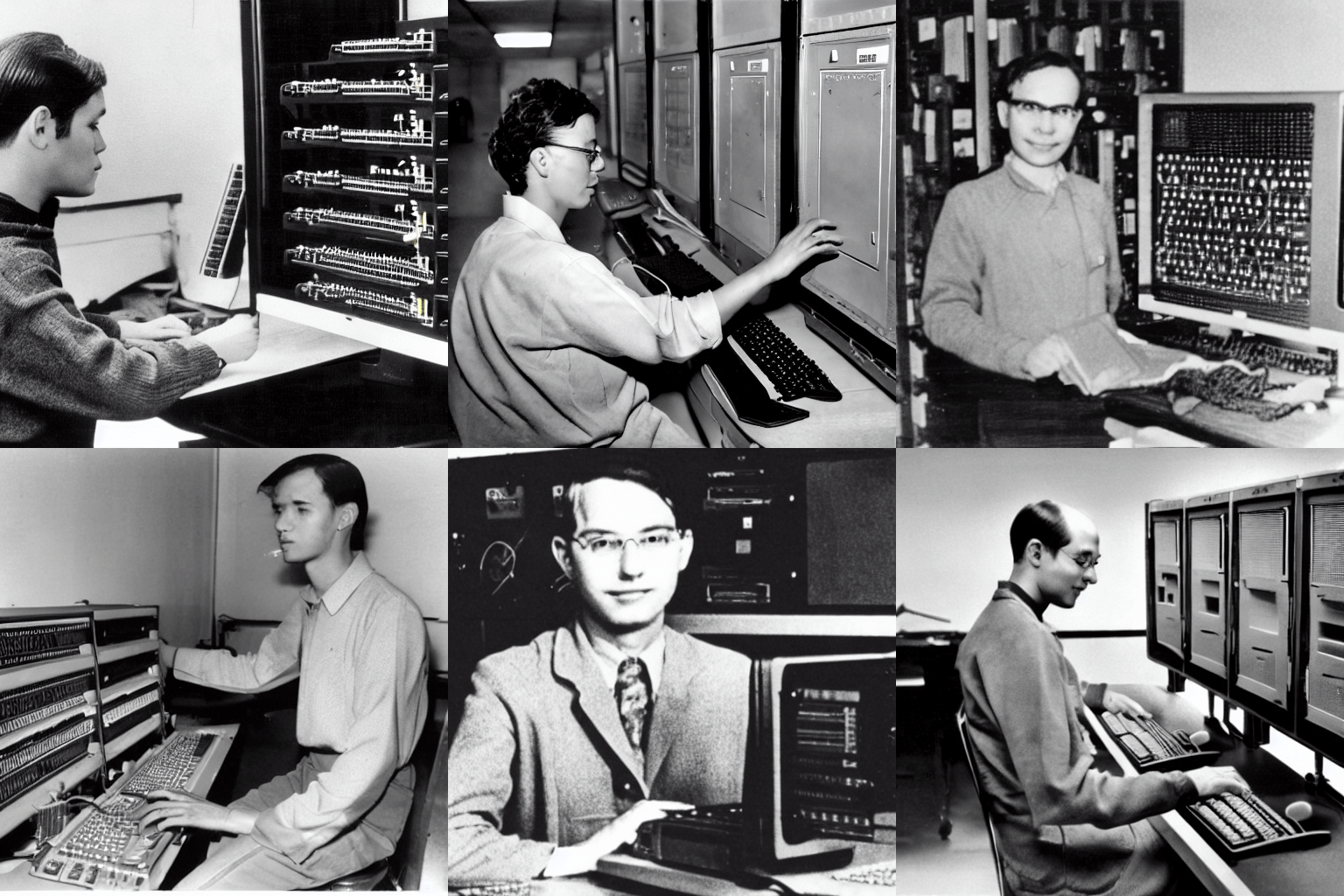}
    \end{minipage}
    \caption{Left: A random sample of the top 100 Google Images results for \emph{``A photo of a computer programmer for the ENIAC''}. Right: Sample generated images by SD 1.4 for the same prompt.}
    \label{fig:query_contextual}
\end{figure}

\noindent \textbf{MPR is prompt contextual.} MPR's flexibility extends to evaluating prompt-specific representational requirements, as demonstrated through our analysis of the prompt \emph{"A photo of a computer programmer for the ENIAC."} It is well-known that early computer programming pioneers were predominantly women, with six women programmers specifically responsible for the ENIAC \cite{bartik2013pioneer,kleiman2022proving}. Given this historical context, accurate image generation should exclusively depict women programmers. To establish a reference distribution, we analyzed the top 100 Google image search results for this prompt, which mostly showed women programmers (Fig. \ref{fig:query_contextual}, left).
Table \ref{tab:contextual_query} presents MPR measurements comparing various model generations against two reference points: an equal gender distribution and the historically accurate women-only representation derived from our Google image search query. The results reveal two distinct representational harms: First, the vanilla model strongly over-represents male programmers (Fig. \ref{fig:query_contextual}, right), diverging from both equal representation and historical accuracy. Second, while a fair generative baseline (i.e., FairDiffusion) achieve gender parity, they fall short of capturing the historically appropriate representation of exclusively women programmers. This analysis demonstrates MPR's capability to identify not only deviations from equal representation but also failures to preserve important historical context. The results of other baselines are described in Section \ref{append:eniac} in the supplementary material.

\noindent \textbf{MPR can scale to a large set of attributes.} Balancing representation for each identity combination leads to computational complexity that grows exponentially with the number of attribute categories to account for all possible intersectional groups. In contrast, MPR's function class $\mathcal{C}$ can efficiently process attribute vectors of arbitrary dimensionality, enabling the measurement of complex, high-dimensional representational patterns. This capability is particularly relevant given recent developments like the multifaceted identity embedding framework introduced in \citep{srinivasan2024generalized}. MPR can evaluate representational fairness over such rich embeddings without incurring additional computational overhead, though we note that the specific embeddings from \cite{srinivasan2024generalized} remain proprietary and were not available for our experimental validation.

%% file: CVPR_2025/tables/table1.tex
\begin{table*}[t!]
\scriptsize
\setlength{\tabcolsep}{2pt}
\centering
\caption{\textbf{MPR results for several traits}. Every MPR is obtained with groups of ``Gender'', ``Age'' and ``Race'' and decision trees. The splits highlighted in bold font indicate those used in the decision trees with a depth of 1. The number in parenthesis represents the standard deviation estimated from bootstrapping the generated images.}
\begin{tabular}{@{}llcccccc@{}}
\toprule
                                                         &                                    & Attractive                                                                       & Emotional                   & Exotic                                                                        & Poor                                                                                   & Terrorist                                                                              & Thug                                                                       \\ \midrule \midrule
\multirow{2}{*}{SD v1.4}                                 & DT$(d=1)$                                & 0.67 ($\pm$0.01)                                                                  & 0.46 ($\pm$0.01)             & 0.39 ($\pm$0.01)                                                               & 0.53 ($\pm$0.01)                                                                        & 0.47 ($\pm$0.00)                                                                        & 0.81 ($\pm$0.01)                                                            \\
                                                         & DT$(d=3)$                                & 0.72 ($\pm$0.01)                                                                  & 0.62 ($\pm$0.01)             & 0.59 ($\pm$0.01)                                                               & 0.55 ($\pm$0.01)                                                                        & 0.61 ($\pm$0.00)                                                                        & 0.84 ($\pm$0.01)                                                            \\\midrule 
\multirow{2}{*}{SD v2.1}                                   & DT$(d=1)$                                & 0.66 ($\pm$0.01)                                                                  & 0.52 ($\pm$0.01)             & 0.30 ($\pm$0.01)                                                               & 0.50 ($\pm$0.00)                                                                        & 0.48 ($\pm$0.01)                                                                        & 0.47 ($\pm$0.01)                                                            \\
                                                         & DT$(d=3)$                                & 0.71 ($\pm$0.01)                                                                  & 0.53 ($\pm$0.01)             & 0.41 ($\pm$0.02)                                                               & 0.60 ($\pm$0.01)                                                                        & 0.79 ($\pm$0.01)                                                                        & 0.60 ($\pm$0.01)                                                            \\\midrule
\multirow{2}{*}{SDXL}                                    & DT$(d=1)$                                & 0.73 ($\pm$0.01)                                                                  & 0.51 ($\pm$0.02)             & 0.43 ($\pm$0.01)                                                               & 0.53 ($\pm$0.01)                                                                        & 0.47 ($\pm$0.00)                                                                        & 0.72 ($\pm$0.01)                                                            \\
                                                         & DT$(d=3)$                                & 0.77 ($\pm$0.01)                                                                  & 0.55 ($\pm$0.01)             & 0.56 ($\pm$0.01)                                                               & 0.56 ($\pm$0.01)                                                                        & 0.79 ($\pm$0.01)                                                                        & 0.80 ($\pm$0.01)                                                            \\\midrule \midrule
\multicolumn{2}{c}{\multirow{3}{*}{\begin{tabular}[c]{@{}c@{}}Splits of DT$(d=3)$ \\ SD v1.4\end{tabular}}} & \multirow{3}{*}{\begin{tabular}[c]{@{}c@{}}Old\\ Southeast Asian \\ \textbf{White}\end{tabular}} & \multirow{3}{*}{\begin{tabular}[c]{@{}c@{}}\textbf{Male}\\ White \\ Latino\end{tabular}}& \multirow{3}{*}{\begin{tabular}[c]{@{}c@{}}Old\\ Black \\ \textbf{Indian}\end{tabular}} & \multirow{3}{*}{\begin{tabular}[c]{@{}c@{}}Male\\ \textbf{Old} \\ \textbf{Indian}\end{tabular}} & \multirow{3}{*}{\begin{tabular}[c]{@{}c@{}}\textbf{Male}\\ Black \\ East Asian\end{tabular}} & \multirow{3}{*}{\begin{tabular}[c]{@{}c@{}}Male\\ Old \\ \textbf{Black}\end{tabular}} \\ 
\multicolumn{2}{l}{}                                                                          &                                                                                  &                             &                                                                               &                                                                                        &                                                                                        &                                                                            \\ 
\multicolumn{2}{l}{}                                                                          &                                                                                  &                             &                                                                               &                                                                                        &                                                                                        &                                                                            \\ \midrule
\multicolumn{2}{c}{\multirow{3}{*}{\begin{tabular}[c]{@{}c@{}}Splits of DT$(d=3)$ \\ SD v2.1\end{tabular}}} & \multirow{3}{*}{\begin{tabular}[c]{@{}c@{}}Male\\ Old\\ \textbf{White}\end{tabular}} &\multirow{3}{*}{\begin{tabular}[c]{@{}c@{}}Old\\ East Asian \\ \textbf{White}\end{tabular}} & \multirow{3}{*}{\begin{tabular}[c]{@{}c@{}}Old\\ \textbf{Indian} \\ Latino\end{tabular}} & \multirow{3}{*}{\begin{tabular}[c]{@{}c@{}}\textbf{Old}\\ Indian \\ White\end{tabular}} & \multirow{3}{*}{\begin{tabular}[c]{@{}c@{}}Male\\ Indian \\ \textbf{Middle Eastern}\end{tabular}} & \multirow{3}{*}{\begin{tabular}[c]{@{}c@{}}\textbf{Male}\\ Black\\ Latino\end{tabular}} \\
\multicolumn{2}{l}{}                                                                          &                                                                                  &                             &                                                                               &                                                                                        &                                                                                        &                                                                            \\ 
\multicolumn{2}{l}{}                                                                          &                                                                                  &                             &                                                                               &                                                                                        &                                                                                        &                                                                            \\ \midrule
\multicolumn{2}{c}{\multirow{3}{*}{\begin{tabular}[c]{@{}c@{}}Splits of DT$(d=3)$ \\ SDXL\end{tabular}}} & \multirow{3}{*}{\begin{tabular}[c]{@{}c@{}}Male\\ Old \\ \textbf{White}\end{tabular}} & \multirow{3}{*}{\begin{tabular}[c]{@{}c@{}}Old\\ East Asian \\ \textbf{White}\end{tabular}} & \multirow{3}{*}{\begin{tabular}[c]{@{}c@{}}\textbf{Male}\\ Age \\ White\end{tabular}} & \multirow{3}{*}{\begin{tabular}[c]{@{}c@{}}Male\\ \textbf{Old} \\ Black\end{tabular}} & \multirow{3}{*}{\begin{tabular}[c]{@{}c@{}}\textbf{Male}\\ Indian \\ Middle Eastern\end{tabular}} & \multirow{3}{*}{\begin{tabular}[c]{@{}c@{}}Male\\ Old \\ \textbf{Black}\end{tabular}} \\
\multicolumn{2}{l}{}                                                                          &                                                                                  &                             &                                                                               &                                                                                        &                                                                                        &                                                                            \\
\multicolumn{2}{l}{}                                                                          &                                                                                  &                             &                                                                               &                                                                                        &                                                                                        &                                                                            \\  \bottomrule
\end{tabular}
\label{tab:mpr_trait}
\end{table*}

%% file: CVPR_2025/tables/table2.tex
\begin{table}[t!]
\centering
\scriptsize
\caption{\small {\bf MPRs for disability representation.} WC is short for a wheelchair user.}
\begin{tabular}{llcl}
\toprule
                         &     & MPR      & \multicolumn{1}{c}{Splits} \\ \midrule \midrule
\multirow{2}{*}{SD v1.4} & DT (\scriptsize $d\tteq 1$)& 0.53 {\scriptsize($\pm$0.01)} & White                      \\
                         & DT (\scriptsize $d\tteq 2$)& 0.54 {\scriptsize($\pm$0.02)} & WC / White                 \\ \midrule
\multirow{2}{*}{SD v2.1}   & DT (\scriptsize $d\tteq 1$)& 0.50 {\scriptsize($\pm$0.01)} & WC                         \\
                         & DT (\scriptsize $d\tteq 2$)& 0.54 {\scriptsize($\pm$0.01)} & WC / Asian                 \\ \midrule
\multirow{2}{*}{SDXL}    & DT (\scriptsize $d\tteq 1$)& 0.55 {\scriptsize($\pm$0.01)} & WC                         \\
                         & DT (\scriptsize $d\tteq 2$)& 0.61 {\scriptsize($\pm$0.01)} & WC / Black                \\
                 \bottomrule
\end{tabular}
\label{tab:mpr_disability}
\end{table}


%% file: CVPR_2025/tables/table3.tex
\begin{table}[t]
\centering
\scriptsize
\caption{\small {\bf Query-Contextual MPR.} Balanced (left) and contextual (right) reference set. MPR is calculated with the decision trees with a depth of 1 and ``gender'' as a sensitive attribute.}
\vspace{-.1in}
\begin{tabular}{@{}lll@{}}
\toprule
Reference     & Equal          & Google Images  \\ \midrule \midrule
SD v1.4       & 0.40 {\scriptsize($\pm$ 0.01)} & 0.69 {\scriptsize($\pm$ 0.01)} \\
FairDiffusion \cite{friedrich2023fair} & 0.03 {\scriptsize($\pm$ 0.01)} & 0.32 {\scriptsize($\pm$ 0.01)} \\
\bottomrule
\end{tabular}
\label{tab:contextual_query}
\vspace{-.2in}
\end{table}


%% file: CVPR_2025/sec/5_method.tex
\section{Fine-tuning T2I models for MPR}\label{sec:intervention_methods}

We propose a fine-tuning method to reduce representational harm in T2I models in terms of MPR. Given a reference dataset, we can improve MPR of a T2I model by utilizing Equation \eqref{emp_mpr_def} as an objective function. However, minimizing this function directly via gradient descent is non-trivial for two main reasons: (1) estimating MPR requires generating a large number of images, which would result in significant computational overhead if done at every update, and (2) conducting gradient descent is challenging because of the supremum operation in \eqref{emp_mpr_def}. At each iteration, we can compute the $c^*$ which maximizes MPR for a specific set of generated images using Props. \ref{linear} and \ref{dt}. However, we observed that just updating the model to reduce MPR for the current $c^*$ may not reduce it over the full set of functions (the supremum). For example, the model may first optimize gender representation, then switch to race representation, and in doing so lose the progress it gained on gender representation. To address the aforementioned issues, we store generated images and the $c^*$ functions into buffers at each update. 
Then, we approximate MPR in every update step using the buffers together with the newly generated images and computed $c^*$. By doing so, we can avoid the need to generate a large number of images per iteration and address the supremum. The buffers are updated in a First-In-First-Out (FIFO) manner, constrained by predefined sizes. Additionally, minimizing only Equation \eqref{emp_mpr_def} could cause the finetuned model $\theta_t$ at iteration $t$ to deviate excessively from the pre-trained model $\theta_0$, potentially degrading image generation quality. To prevent this, we incorporate regularization terms based on the CLIP and DINO models, as proposed in \cite{shen2023finetuning}. Our final objective function is $\mathcal{L}_\text{obj}(\theta_t) = \mathcal{L}_\text{MPR}(\theta_t) + \lambda \mathcal{L}_\text{img}(\theta_t)$, where
\begin{align}
    &\mathcal{L}_\text{MPR}(\theta_t) = \sum_{c\in\hat{\mathcal{C}}}\big| \frac{1}{k}\sum_{x_i\in \hat{G}} c(x_i) - \frac{1}{m} \sum_{x_j\in D} c(x_j)\big|, \nonumber  \\
    & \mathcal{L}_\text{img}(\theta_t) =\frac{1}{N}\sum_{i=1}^N \bigg[ (1-\text{cos}(\text{CLIP}(x_i), \text{CLIP}(o_i)) \nonumber \\ &\qquad\qquad\qquad+ (1-\text{cos}(\text{DINO}(x_i), \text{DINO}(o_i)))\bigg]. \nonumber 
\end{align}
\normalsize
Here, $\lambda$ is a hyperparameter for the regularization terms, $x_i, o_i$ are samples generated from $\theta_t, \theta_0$ respectively, and $\hat{G}$ and $\hat{\mathcal{C}}$ contain generated images and $c$ functions obtained from the current and previous updates. We adopt the gradient-adjusting technique \cite{shen2023finetuning} to prevent gradient explosion and utilize LoRA \cite{hu2021lora} for computationally efficient learning. Our fine-tuning method is detailed in Alg. \ref{alg:ours}.

\paragraph{Performance comparison.} We validate the effectiveness of our method by examining its ability to improve MPR values for the occupation representation. Specifically, we applied our method and existing baseline methods for seven professions (chef, therapist, taxi driver, flight attendant, housekeeper, nurse) that were qualitatively assessed in \cite{bianchi2023easily}. We evaluated all methods in terms of representational harm and image quality using two types of MPR values and the CLIP similarity between the occupation prompts and the generated images. To ensure a fair comparison, all baseline methods were also adapted to use statistics derived from the reference dataset.

Table \ref{tab:performance_comparison} demonstrates that methods like FairDiffusion and UCE, which focus solely on a single group attribute (gender), perform poorly when the MPR metric considers multiple attributes (gender, age, and race). Although ITI-GEN achieves reasonable performance on both MPR metrics, its lack of consideration for image quality results in slightly degraded CLIP scores. In contrast, our method achieves the best performance across both types of MPR while maintaining robust CLIP scores, ensuring high image quality. The details of our method implementation are provided in Section \ref{appen:implementation_detail} in the supplementary material.

\begin{table}[t!]

\centering
\scriptsize
\caption{\small {\bf Performance comparison.} (G) and (GAR) indicate the group attributes considered: gender, or gender, age, and race, respectively. The reported MPR values and CLIP scores are averages across seven professions (chef, therapist, taxi driver, flight attendant, housekeeper, nurse).}
\label{tab:performance_comparison}
\begin{tabular}{llll}
\toprule
              & MPR \scriptsize{(G)}         & MPR \scriptsize{(GAR)}       & CLIP score      \\ \midrule \midrule
Vanilla       & 1.18 \scriptsize{($\pm$0.02)} & 1.89 \scriptsize{($\pm$0.02)} & 0.31 \scriptsize{($\pm$0.01)} \\ \midrule
Entigen \cite{bansal2022well}      & 1.05 \scriptsize{($\pm$0.03)} & 1.61 \scriptsize{($\pm$0.03)} & 0.30 \scriptsize{($\pm$0.01)} \\
FairDiffusion \cite{friedrich2023fair} & 0.27 \scriptsize{($\pm$0.04)} & 1.32 \scriptsize{($\pm$0.02)} & 0.31 \scriptsize{($\pm$0.02)} \\
UCE \cite{gandikota2024unified}          & 0.80 \scriptsize{($\pm$0.05)} & 1.56 \scriptsize{($\pm$0.04)} & 0.31 \scriptsize{($\pm$0.01)} \\
ITI-GEN \cite{zhang2023iti}      & 0.13 \scriptsize{($\pm$0.04)} & 0.44 \scriptsize{($\pm$0.02)} & 0.28 \scriptsize{($\pm$0.03)} \\ \midrule
Ours          & \textbf{0.08} \scriptsize{($\pm$0.05)} & \textbf{0.25} \scriptsize{($\pm$0.03)} & 0.30 \scriptsize{($\pm$0.01)} \\
\bottomrule
\end{tabular}
\end{table}








%% file: CVPR_2025/sec/9_conclusion.tex
\section{Concluding Remarks and Limitations}\label{sec:conclusion}

We introduced Multi-Group Proportional Representation (MPR) as a comprehensive framework for measuring representational harm in text-to-image generative models. Unlike existing methods that often focus on a single or limited number of sensitive attributes, MPR supports harm measurement with respect to arbitrary sensitive attributes and their intersections. Through extensive experiments, we quantitatively measured significant representational disparities across professions, traits, and disabilities that prior works addressed only qualitatively. Additionally, we demonstrate that using MPR as an optimization objective during model fine-tuning effectively reduces intersectional disparities while maintaining image quality. While important challenges remain, including the need for robust attribute estimation methods and context-appropriate reference distributions, our work provides a systematic approach to measuring and mitigating representational harms. As text-to-image models become increasingly prevalent, frameworks like MPR will be crucial to ensuring that these systems serve and represent all members of society equitably.


\newpage 

\section*{Acknowledgments}
We would like to thank Himabindu Lakkaraju for her helpful discussions and valuable insights during the preparation of this paper.

This work was supported in part by the National Research Foundation of Korea (NRF) grant [RS-2021-NR059237] and by Institute of Information \& communications Technology Planning \& Evaluation (IITP) grants [RS-2021-II211343, RS-2021-II212068, RS-2022-II220113, RS-2022-II220959] funded by the Korean government (MSIT). It was also supported by AOARD Grant No. FA2386-25-1-4015, Brain Korea 21 Plus Project and Hyundai Motor Chung Mong-Koo Foundation. AO is supported by the National Science Foundation Graduate Research Fellowship under Grant No. DGE-2140743. This material is based upon work supported by the National Science Foundation under awards CIF-2231707,  CIF-2312667, and FAI-2040880. 


%% file: CVPR_2025/sec/X_suppl.tex

\onecolumn 
\newtheorem{lemma}{Lemma}
\appendix
\setcounter{page}{1}
\section*{Supplementary Material}

In this supplementary material, we provide detailed proofs, additional experimental results, and implementation details that complement our main paper. The appendix is organized as follows:

Section \ref{sec:appendix_limitation} outlines the limitations of our approach, discussing challenges related to attribute estimation, reference population collection, and prompt construction.
Section \ref{sec:appendix_proofs} presents complete mathematical proofs for all propositions stated in Section \ref{sec:MPR_metric} of the main paper. In particular, we establish the Generalization bound for the Multi-group proportional metric, stated in Proposition \ref{gen}, and we characterize MPR for the class of bounded linear functions and decision trees, stated in Propositions \ref{linear} and \ref{dt} respectively. Section \ref{appen:implementation_detail} provides comprehensive implementation details, including our methodology for group label creation and the implementation specifics of baseline and fine-tuning methods.
Section \ref{appen:additional_experiments} contains additional experimental results, including

\begin{itemize}
\item[i.] MPR results for trait representation across multiple diffusion models.
\item[ii.] Analysis of empirical vs. true MPR gaps across different function classes.
\item[iii.] Additional baseline method results for the ENIAC programmer case.
\item[iv.] Supplementary qualitative results.
\end{itemize}


Section \ref{appen:practical_guidelines} discusses practical considerations for selecting parameters $k$ and $m$ in the MPR framework, including guidance for real-world applications and empirical validation.


\section{Limitations}\label{sec:appendix_limitation}

While MPR represents a significant advance in measuring and mitigating representational bias in text-to-image systems, there are several important limitations to our approach that warrant careful consideration. One fundamental challenge lies in our reliance on automated systems to estimate demographic attributes such as gender, race, and disability status. These estimation methods may not only perpetuate harmful categorization practices but could also contain inherent biases themselves. The binary classification of complex social identities oversimplifies human diversity and risks reinforcing problematic societal categorizations. This is particularly concerning for attributes like disability status, which are both technically challenging to detect and ethically sensitive to classify.

The selection of appropriate reference distributions presents another significant challenge. Determining what constitutes appropriate representation is inherently complex and context-dependent, with different stakeholders potentially holding varying views on fair representation. Historical data used for reference distributions may contain existing societal biases, creating a tension between reflecting historical accuracy and promoting more equitable representation. Additionally, there is a notable scarcity of balanced, intersectional datasets across many domains, limiting our ability to establish comprehensive benchmarks for fair representation. We note that FairFace dataset is one of the most comprehensive facial image datasets, which is debiased toward gender, age and race. While some of our MPR evaluations rely on the FairFace dataset along gender, age, and race axes, they are merely examples of use cases for MPR and do not imply a limitation of MPR. Furthermore, we believe that the cost and effort required to create robust reference distributions are ultimately inevitable for measuring and mitigating biases in image generation models. The key contribution of our work lies in proposing a flexible measurement framework that can adapt to such advanced reference distributions as they are developed.

Our current evaluation methodology also has limitations in terms of prompt engineering and assessment. The focus on simple prompt structures (e.g., ``\emph{a photo of {concept}}'') may not adequately reflect the complexity and variety of real-world usage. More diverse prompt templates are needed to better represent natural language variation and capture cultural or contextual nuances. The challenge of evaluating abstract concepts and scaling across multiple languages and cultural contexts remains significant.

From a methodological perspective, we face several technical constraints. There exists a fundamental trade-off between function class complexity and sample size requirements, affecting the practical applicability of our approach. The computational costs of evaluating multiple intersectional categories simultaneously can be substantial, and maintaining image quality while optimizing for representational fairness presents ongoing challenges. Our current framework also has limited ability to capture temporal or contextual aspects of representation.

Looking toward future work, these limitations highlight several critical directions for research. We need to develop more nuanced and ethical approaches to attribute detection, create context-aware reference distribution frameworks, and expand to more complex prompt structures and evaluation scenarios. Improving computational efficiency and deepening engagement with affected communities will be crucial. Furthermore, integration with other fairness metrics and evaluation frameworks could provide more comprehensive assessments of representational fairness.

These challenges underscore the complexity of measuring and promoting fairness in generative AI systems. As these technologies continue to evolve and their societal impact grows, addressing these limitations will be essential for developing more equitable and responsible AI systems that truly serve all members of society.

\input{CVPR_2025/sec/X1_proofs}
\input{CVPR_2025/sec/X2_implementation_details}
\input{CVPR_2025/sec/X3_additional_results}
\input{CVPR_2025/sec/X4_sample_sizes}

\begin{figure}[t!]
    \centering
    \includegraphics[width=0.7\textwidth]{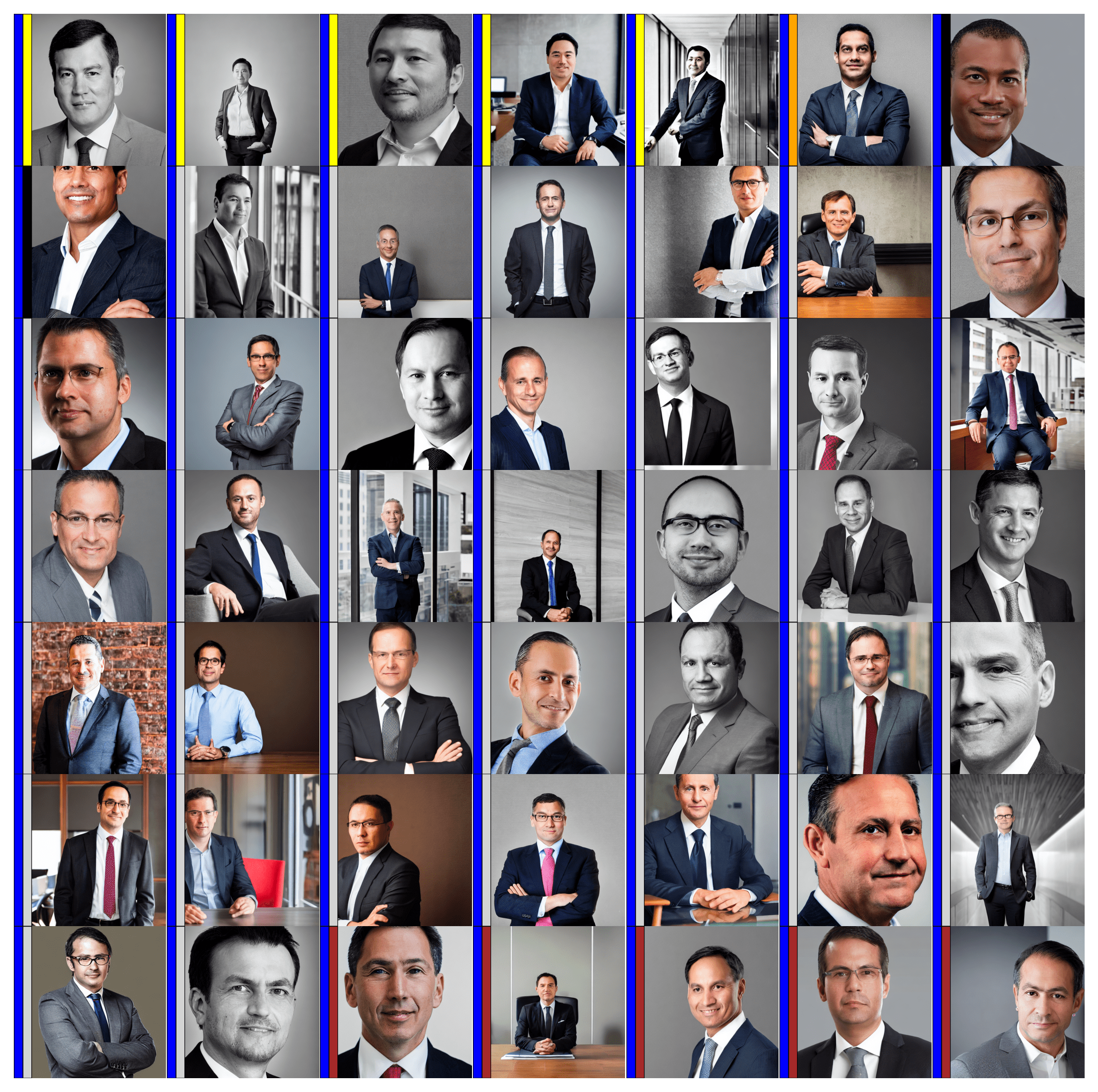}
    \caption{\small \textbf{Images generated from the ``CEO'' prompt using the vanilla SD v1.4 model.} The leftmost and second color bars for each image represent the estimated gender and race, respectively. Gender is indicated by blue for males and red for females. Race is represented by the following colors: grey for White, black for Black, yellow for East Asian, green for Southeast Asian, orange for Indian, brown for Latino/Hispanic, and purple for Middle Eastern.}
    \label{fig:scratch_example}
\end{figure}
\begin{figure}[t!]
    \centering
    \includegraphics[width=0.7\textwidth]{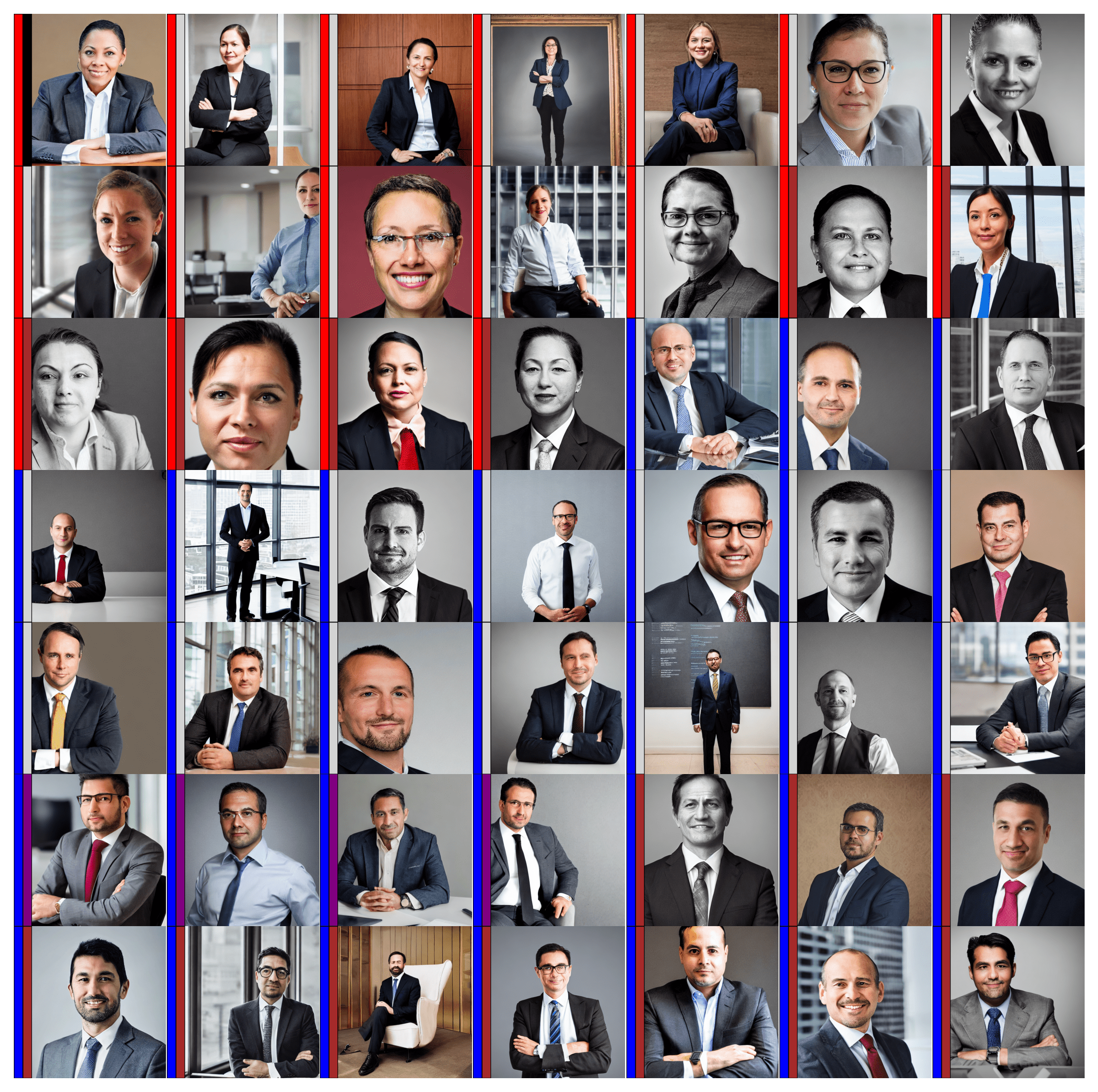}
    \caption{\small \textbf{Images generated from the ``CEO'' prompts using FairDiffusion.} The color details follow the same scheme as Fig. \ref{fig:scratch_example}.}
    \label{fig:fairdiffusion_example}
\end{figure}
\begin{figure}[t!]
    \centering
    \includegraphics[width=0.7\textwidth]{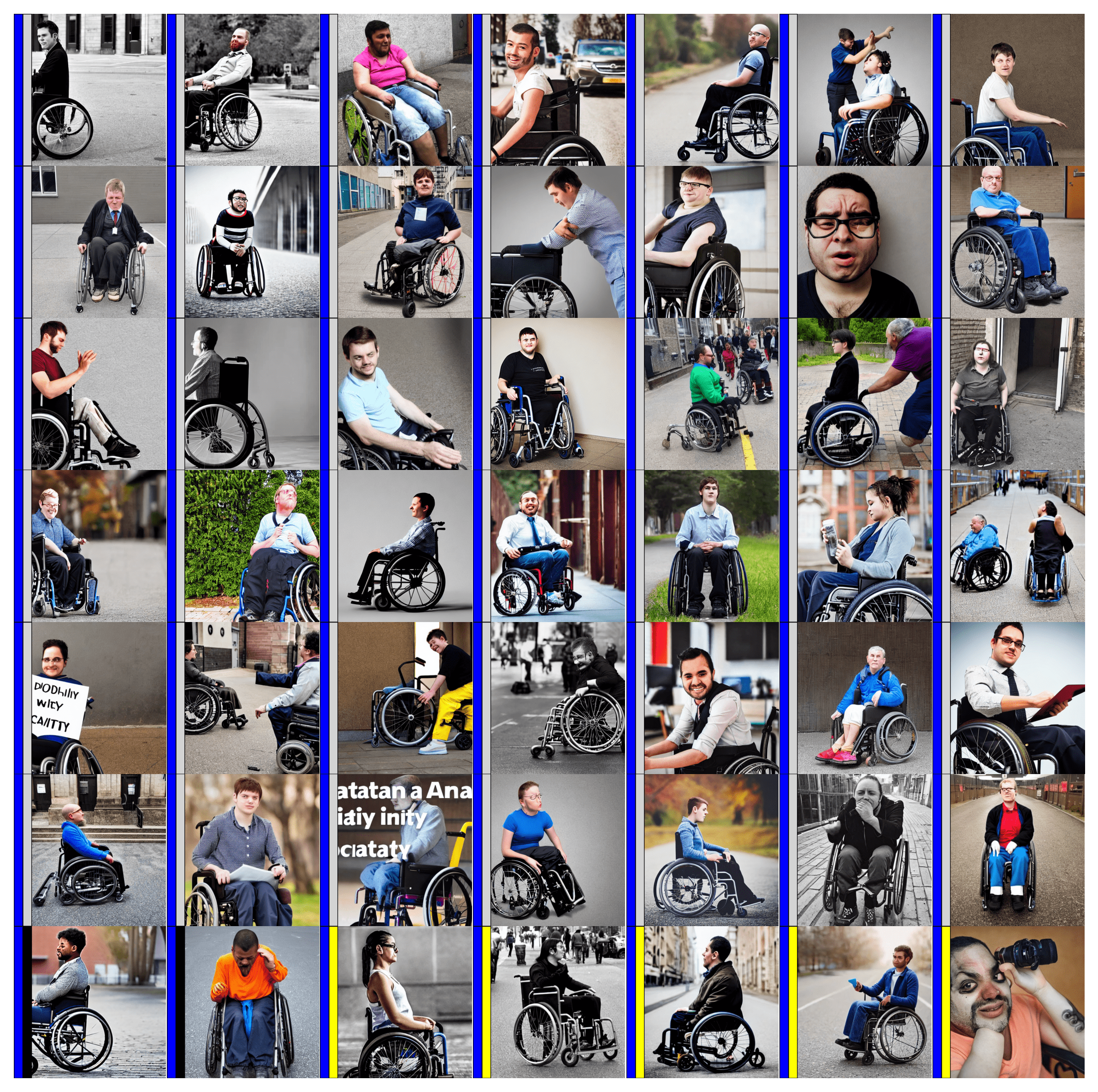}
    \caption{\small \textbf{Images generated from the ``disability'' prompts using the vanilla SD v1.4.} The leftmost and second color bars for each image represent the estimated presence of a wheelchair and race, respectively. The presence of a wheelchair is indicated by blue for wheelchair users and red for non-wheelchair users. Race is represented by the following colors: grey for White, black for Black, yellow for Latino/Hispanic, and purple for Asian.}
    \label{fig:disability_example}
\end{figure}
%
%

%% file: CVPR_2025/sec/X1_proofs.tex
\section{Proofs}\label{sec:appendix_proofs}
\textbf{Proposition~\ref{gen} restated:} For any given prompt $q$, let $\widehat{G}_q$ and $\widehat{R}$ be the sets of generated and reference samples used to approximate $G_q$ and $R$, respectively.
Then, for any bounded class of functions $\mathcal{C}$ and any $\delta>0$,
\begin{align}
    &\left|\mathrm{MPR}(\mathcal{C},\widehat{G}_q,\widehat{R})-\mathrm{MPR}(\mathcal{C},G_q,R)\right|\leq 2\mathcal{R}_{\widehat{G}_q}(\mathcal{C}) + 2\mathcal{R}_{\widehat{R}}(\mathcal{C}) + B\sqrt{\frac{\log\left( \frac{2}{\delta}\right)}{2(k+m)}} \label{eq:prop1}
\end{align}
with probability at least $1 - \delta$, where $B=\sup_{\substack{c\in\mathcal{C}\\X\neq X'}}|c(X)-c(X')|$ and $\mathcal{R}_\mathcal{A}(\mathcal{C})$ is the Rademacher complexity of $\mathcal{C}$ relative to the set $\mathcal{A}$.

\begin{proof}
Define $f(c):=\mathrm{MPR}(\mathcal{C},G_q,R)=\mathbb{E}_{G_q}[c(X_g)]-\mathbb{E}_R[c(X_r)]$ and $\hat{f}(c):=\mathrm{MPR}(\mathcal{C},\widehat{G}_q,\widehat{R})=\frac{1}{k}\sum_{i=1}^kc(x_i^g)-\frac{1}{m}\sum_{j=1}^mc(x_j^r)$. To start, note that,
\begin{align}\label{mid}
    \Big|\sup_{c\in\mathcal{C}} \hat{f}(c)-\sup_{c\in\mathcal{C}} f(c)\Big|&\leq\sup_{c\in\mathcal{C}} \Big|\hat{f}(c)- f(c)\Big|
\end{align}
which implies,
\begin{align}
    \mathbb{P}\Big(\Big|\sup_{c\in\mathcal{C}} \hat{f}(c)-\sup_{c\in\mathcal{C}} f(c)\Big|\geq\epsilon\Big)&\leq\mathbb{P}\Big(\sup_{c\in\mathcal{C}} \Big|\hat{f}(c)- f(c)\Big|\geq\epsilon\Big)\\
    &\leq\mathbb{P}\Big(\sup_{c\in\mathcal{C}} \left(\hat{f}(c)- f(c)\right)\geq\frac{\epsilon}{2}\Big)+\mathbb{P}\Big(\sup_{c\in\mathcal{C}}\left( f(c)-\hat{f}(c)\right)\geq\frac{\epsilon}{2}\Big)\label{mid0}
\end{align}
Next, we analyze each probability term in \eqref{mid0}. Define $g(x_1^g,\dotsc,x_k^g,x_1^r,\dotsc,x_m^r)=\sup_{c\in\mathcal{C}}(\hat{f}(c)-f(c))$. Let $\{z\}_{i=1}^{k+m}=\{\{x_i^g\}_{i=1}^n,\{x_i^r\}_{i=1}^m\}$ be the concatenation of $\mathcal{G}$ and $\mathcal{D}$. For any $i\in\{1,\dotsc,k+m\}$,
\begin{align}
    \big|g(z_1,\dotsc,z_{i-1},z_i,z_{i+1},\dotsc,z_{k+m})\nonumber-g(z_1,\dotsc,z_{i-1},z'_i,z_{i+1},\dotsc,z_{k+m})\Big|\nonumber &\leq\sup_{c\in\mathcal{C}}\frac{1}{\min\{m,k\}}\Big|c(z_i)-c(z_i')\Big|\label{mid1}\\
    &=\frac{B}{\min\{m,k\}},
\end{align}
where \eqref{mid1} follows from \eqref{mid}. Then, from McDiarmid's inequality \cite{mcdiarmid1989method,boucheron2013concentration}, we have,
\begin{align}
    \mathbb{P}\Big(\sup_{c\in\mathcal{C}}(\hat{f}(c)-f(c))\geq\epsilon + \mathbb{E}[\sup_{c\in\mathcal{C}}(\hat{f}(c)-f(c))]\Big)\nonumber
    \leq e^{-\frac{2\epsilon^2(m+k)}{B^2}}.
\end{align}

Now, we bound the expectation of the right-hand side inside the probability with a symmetrization argument. Indeed, Let $\sigma_i$ be independent Rademacher random variables (uniformly distributed on $\{-1, 1\}$.
\begin{align}
\mathbb{E}\big[\sup_{c\in\mathcal{C}}(\hat{f}(c)- f(c))\big] &= \mathbb{E}\bigg[\sup_{c\in\mathcal{C}}\left(\mathrm{MPR}(\mathcal{C},\widehat{G}_{q},\widehat{R})-\mathbb{E}\big[\mathrm{MPR}(\mathcal{C},\widehat{G}_q',\widehat{R}')\big]\right)\bigg]\\
& = \mathbb{E}\bigg[\sup_{c\in\mathcal{C}} \left(\mathrm{MPR}(\mathcal{C},\widehat{G}_q,\widehat{R}\right)-\mathbb{E}\big[\mathrm{MPR}(\mathcal{C},\widehat{G}_q',\widehat{R}') | \widehat{G}_q, \widehat{R}\big]\bigg] \\
& = \mathbb{E}\bigg[\sup_{c\in\mathcal{C}} \mathbb{E} \big[\mathrm{MPR}(\mathcal{C},\widehat{G}_q,\widehat{R})-\mathrm{MPR}(\mathcal{C},\widehat{G}_q',\widehat{R}') | \widehat{G}_q, \widehat{R}\big]\bigg] \\
& \leq \mathbb{E}\bigg[\mathbb{E} \big[\sup_{c\in\mathcal{C}}(\mathrm{MPR}(\mathcal{C},\widehat{G}_q,\widehat{R})-\mathrm{MPR}(\mathcal{C},\widehat{G}_q',\widehat{R}')) | \widehat{G}_q, \widehat{R}\big]\bigg] \\
& = \mathbb{E} \bigg[\sup_{c\in\mathcal{C}}\mathrm{MPR}(\mathcal{C},\widehat{G}_q,\widehat{R})-\mathrm{MPR}(\mathcal{C},\widehat{G}_q',\widehat{R}')\bigg] \\
& = \mathbb{E} \sup_{c\in\mathcal{C}} \left[ \frac{1}{k}\sum_{i=1}^k c(x_i) - \frac{1}{m} \sum_{i=1}^m c(x_i)- \frac{1}{k} \sum_{i=1}^k c(\tilde{x}_i)+ \frac{1}{m} \sum_{i=1}^mc(\tilde{x}_i)\right]\\
& = \mathbb{E} \sup_{c\in\mathcal{C}} \bigg[ \frac{1}{k}\sum_{i=1}^{k} \sigma( c(x_i)-c(\tilde{x_i}) ) - \frac{1}{m} \sum_{i=k+1}^{k+m}\sigma ( c(x_i)-c(\tilde{x_i})) \bigg] \\
&\leq \mathbb{E} \left[\sup_{c\in\mathcal{C}} \frac{1}{k}\sum_{i=1}^{k} \sigma c(x_i)+ \sup_{c\in\mathcal{C}}  \frac{1}{k}\sum_{i=1}^{k} \!\!\sigma c(\tilde{x_i}) + \sup_{c\in\mathcal{C}} \sum_{i=k+1}^{k+m} -\sigma c(\tilde{x_i}) + \sup_{c\in\mathcal{C}} \sum_{i=k+1}^{k+m} - \sigma c(\tilde{x_i})\right]\\
& = 2 \mathbb{E} \sup_{c\in\mathcal{C}}  \frac{1}{k} \sum_{i=1}^{k} \sigma c(x_i) +  2 \mathbb{E} \sup_{c\in\mathcal{C}}  \frac{1}{k} \sum_{i=n+1}^{k+m} \sigma c(x_i)\\
&=2\mathcal{R}_{\widehat{\mathcal{G}}_q}(\mathcal{C}) + 2\mathcal{R}_{\widehat{R}}(\mathcal{C}).
\end{align}
where, by an abuse of notation, we denote by $\widehat{G}_q',\widehat{R}'$ the symmetric i.i.d. copies of $G_q,R$ and $\mathcal{G}_q'$ and $\mathcal{R}'$ denote the sample sets (with samples $\tilde{x}_i$) obtained by the symmetric random variables which empirical distributions given by $\widehat{G}_q'$ and $\widehat{R}'$, respectively.

Now, by choosing $\frac{\epsilon}{2}=t+2\mathcal{R}_{\hat{\mathcal{G}}_q}(\mathcal{C}) + 2\mathcal{R}_{\hat{\mathcal{R}}}(\mathcal{C})$ and $t = B\sqrt{\frac{\ln\frac{1}{\delta}}{2(k+m)}}$, from \eqref{mid0}, we have,
\begin{equation}
    \mathbb{P}\Big(\left|\mathrm{MPR}(\mathcal{C},\widehat{G}_q,\widehat{R})-\mathrm{MPR}(\mathcal{C},G_q,R)  \right|\geq\epsilon\Big)\leq 2e^{-\frac{2\epsilon^2(k+m)}{B^2}}.
\end{equation}
By taking $1-\delta=1-2e^{-\frac{2\epsilon^2(k+m)}{B^2}}$, we prove that,
\begin{align}
    \left|\mathrm{MPR}(\mathcal{C},\widehat{G}_q,\widehat{R})-\mathrm{MPR}(\mathcal{C},G_q,R)  \right|
    \leq2\mathcal{R}_{\hat{\mathcal{G}}_q}(\mathcal{C}) + 2\mathcal{R}_{\hat{\mathcal{R}}}(\mathcal{C})+B\sqrt{\frac{\ln\frac{1}{\delta}}{2(k+m)}}
\end{align}
with probability at least $1-\delta$.
\end{proof}

\textbf{Proposition~\ref{gen_prompt} restated:}  Let $P$ denote the distribution of prompts, and $\{q_1,\dotsc,q_N\}$ denote a set of independent prompts sampled from $P$. Then,
\begin{align}
    \mathbb{P}\left(\left|\frac{1}{N}\sum_{i=1}^N \mathrm{MPR}(\mathcal{C},\widehat{G}_{q_i},\widehat{R}_{q_i})-\mathbb{E}_{Q\sim P}[\mathrm{MPR}(\mathcal{C},G_{Q},R_{Q})]\right|\geq\epsilon\right)\nonumber
    \leq \exp\left(-\frac{\epsilon^2N}{8}\right)+\exp\left(-\frac{2(m+n)}{B^2}\left(\frac{\epsilon}{2}-2\lambda\right)\right)
\end{align}
where $m$ and $n$ are the numbers of generated and reference samples used in the calculation of $\mathrm{MPR}(\mathcal{C},\widehat{G}_{q_i},\widehat{R}_{q_i})$, $Q$ is the random variable representing a prompt, $Q\sim P$, and $\lambda=\sup_{Q\sim P}\mathcal{R}_{\mathcal{G_Q}}+\mathcal{R}_{\mathcal{D_Q}}$. 

\begin{proof}

First, we decompose the difference using the triangle inequality:
\begin{align*}
    \left|\frac{1}{N}\sum_{i=1}^N \text{MPR}(C, \hat{G}_{q_i}, \hat{R}_{q_i}) - \mathbb{E}_{Q\sim P}[\text{MPR}(C, G_Q, R_Q)]\right| & \leq \left|\frac{1}{N}\sum_{i=1}^N \text{MPR}(C, \hat{G}_{q_i}, \hat{R}_{q_i}) - \frac{1}{N}\sum_{i=1}^N \text{MPR}(C, G_{q_i}, R_{q_i})\right| \\
    & \quad + \left|\frac{1}{N}\sum_{i=1}^N \text{MPR}(C, G_{q_i}, R_{q_i}) - \mathbb{E}_{Q\sim P}[\text{MPR}(C, G_Q, R_Q)]\right|
\end{align*}

We start by bounding the first term. From Proposition \ref{gen}, for any fixed $q$:
\begin{equation*}
    \mathbb{P}(|\text{MPR}(C, \hat{G}_q, \hat{R}_q) - \text{MPR}(C, G_q, R_q)| \geq \epsilon/2) \leq \exp\left(-\frac{2(m+n)(\epsilon/2 - 2\lambda)}{B^2}\right)
\end{equation*}

where $\lambda = \sup_{Q\sim P} (R_{G_Q}(C) + R_{R_Q}(C))$. Therefore:
\begin{align*}
    \mathbb{P}\left(\left|\frac{1}{N}\sum_{i=1}^N [\text{MPR}(C, \hat{G}_{q_i}, \hat{R}_{q_i}) - \text{MPR}(C, G_{q_i}, R_{q_i})]\right| \geq \epsilon/2\right) 
     \leq \exp\left(-\frac{2(m+n)(\epsilon/2 - 2\lambda)}{B^2}\right)
\end{align*}

Now, for the second term, we use Hoeffding's inequality since $\text{MPR}(C, G_q, R_q)$ is bounded in $[0,1]$ and the prompts are sampled i.i.d:
\begin{align*}
     \mathbb{P}\left(\left|\frac{1}{N}\sum_{i=1}^N \text{MPR}(C, G_{q_i}, R_{q_i}) - \mathbb{E}_{Q\sim P}[\text{MPR}(C, G_Q, R_Q)]\right| \geq \epsilon/2\right) 
     \leq \exp\left(-\frac{N\epsilon^2}{8}\right)
\end{align*}

To finish, by using a union bound:
\begin{align*}
    \mathbb{P}\left(\left|\frac{1}{N}\sum_{i=1}^N \text{MPR}(C, \hat{G}_{q_i}, \hat{R}_{q_i}) - \mathbb{E}_{Q\sim P}[\text{MPR}(C, G_Q, R_Q)]\right| \geq \epsilon\right) 
    & \leq \mathbb{P}(\text{First term} \geq \epsilon/2) + \mathbb{P}(\text{Second term} \geq \epsilon/2) \\
    & \leq \exp\left(-\frac{\epsilon^2N}{8}\right) + \exp\left(-\frac{2(m+n)(\epsilon/2 - 2\lambda)}{B^2}\right)
\end{align*}

\end{proof}

\begin{remark}
The bound in Proposition 2 can be tightened if an estimate of the variance of $\text{MPR}(C, G_q, R)$ across prompts is available. By applying Bernstein's inequality instead of Hoeffding's inequality for the prompt sampling error, we obtain:
\begin{align}
P\left(\left|\frac{1}{N}\sum_{i=1}^{N}\text{MPR}(C, \hat{G}_{q_i}, \hat{R}) - \mathbb{E}_{Q\sim P}[\text{MPR}(C, G_Q, R)]\right| \geq \varepsilon \right) &\leq 2\exp\left(-\frac{N\varepsilon^2}{8\sigma^2 + 4B\varepsilon/3}\right) \\ 
&+ 2\exp\left(-\frac{2(k+m)(\varepsilon/4 - 2\lambda)^2}{B^2}\right)
\end{align}
where $\sigma^2 = \mathbb{E}_{Q\sim P}[(\text{MPR}(C, G_Q, R) - \mathbb{E}_{Q\sim P}[\text{MPR}(C, G_Q, R)])^2]$ is the variance of $\text{MPR}(C, G_q, R)$ across prompts sampled from $P$. The first term bounds the error from prompt sampling, while the second term bounds the average estimation error across prompts. This variance-aware bound becomes particularly valuable when $\sigma^2$ is small relative to the range $B$, which is often the case for specific prompt categories. In practical applications, $\sigma^2$ can be estimated empirically from the sampled prompts using $\hat{\sigma}^2 = \frac{1}{N-1}\sum_{i=1}^{N}[\text{MPR}(C, \hat{G}_{q_i}, \hat{R}) - \frac{1}{N}\sum_{j=1}^{N}\text{MPR}(C, \hat{G}_{q_j}, \hat{R})]^2$, allowing for adaptive confidence intervals that automatically tighten when MPR values exhibit low variance across the prompt distribution.
\end{remark}


    

\textbf{Proposition~\ref{linear} restated:}
    For the class of bounded linear functions, $\mathcal{C}=\{c:w^Tx\big|w\in\mathbb{R}^d, \|w\| \leq 1\}$, 
    \begin{align}
        \text{MPR}(\mathcal{C},\widehat{P},\widehat{Q}) = \frac{1}{||\tilde{a}^TX||}\tilde{a}^TXX^T\tilde{a},
    \end{align}
    in which, $\tilde{a}\in\mathbb{R}^{k+m}$ has $i$-th entry given by $\tilde{a}_i = \mathbbm{1}_{i \leq k}\frac{1}{k}-\mathbbm{1}_{i>n}\frac{1}{m}$ and $X\in \mathbb{R}^{(k+m)\times d}$ is a matrix (row-wise) concatenated with a set of generated images and the reference dataset.

\begin{proof}
For the class of bounded linear functions, $\mathcal{C}=\{c:w^Tx\big|w\in\mathbb{R}^d, \|w\| \leq 1\}$, 
   \begin{align}
        \text{MPR}(\mathcal{C},\widehat{P},\widehat{Q}) &= \sup_{c\in\mathcal{C}} \big|\frac{1}{k}\sum_{i\in\mathcal{G}} c(x_i^g) - \frac{1}{m}\sum_{j\in\mathcal{D}} c(x_j^r)\big|\\
        &=\sup_{w:\|w\|^2\leq1}\tilde{a}^TXw\label{lin}
\end{align} 
For a fixed $X$ and $\tilde{a}$, $\tilde{a}^TXw$ is maximized when $w$ is aligned along $\tilde{a}^TX$ with the maximum norm allowed. i.e.,
\begin{align}
    \arg\sup_{w:\|w\|^2\leq1}\tilde{a}^TXw=\frac{X^T\tilde{a}}{\|X^T\tilde{a}\|}.
\end{align}
Therefore, from \eqref{lin},
\begin{align}
    \text{MPR}(\mathcal{C},\widehat{P},\widehat{Q})
        &=\sup_{w:\|w\|^2\leq1}\tilde{a}^TXw=\frac{\tilde{a}^TXX^T\tilde{a}}{\|\tilde{a}^TX\|}.
\end{align}
\end{proof}

\textbf{Proposition~\ref{dt} restated:}
    For the class of binary decision trees of depth $\ell\leq n$, where $\mathcal{C}=\{c:\{-1,+1\}^n\rightarrow\{-1,+1\}\}$,
    \begin{align}
        \text{MPR}(\mathcal{C},\widehat{G}_q,\widehat{R}) = \max_{I\subset\{1,2,\dots,n\},|I|=\ell} 2\text{TV}(\tilde{G}_q^I,\tilde{R}^I)
    \end{align}
    where $\tilde{G}_q^I$ and $\tilde{R}^I$ are the marginal distributions corresponding to $\widehat{G}_q$ and $\widehat{R}$ over the attributes in set $I$, and $\text{TV}(A,B)$ denotes the total variation distance between distributions $A$ and $B$.

\begin{proof}
Let $c \in\mathcal{C}=DT^k$ be any decision tree of depth at most $k$, specified by a subset of of attributes $I\subset\{1,\dotsc,d\}$. Each $c\in\mathcal{C}$ corresponds to a specific subset of attributes $I$ and a set of group assignments $c_i\in\{-1,+1\}$ for $i=\{1,\dotsc,2^{|I|}\}$, as it partitions the dataset into $2^{|I|}$ disjoint sets $A_1^I, ..., A_{2^{|I|}}^{I}$, where $|I|=k$. On each $A_i^{I}$, $c$ takes a constant value $c_i=\{-1,+1\}$. Then:

\begin{align*}
\sup_{c\in\mathcal{C}}|E_{{G}_q}[c(X_g)] - E_R[c(X_r)]|&= \sup_{c\in\mathcal{C}}\left|\sum_{i=1}^{2^{|I|}} c_i(G_q(A_i^I) - R(A_i^I))\right| \\
&=\max_{I\subset\{1,\dotsc,d\},|I|=k}\quad\max_{c_1,\dotsc,c_k\in\{-1,+1\}}\left|\sum_{i=1}^k c_i(G_q(A_i^I) - R(A_i^I))\right| 
\end{align*}
For a given set of attributes $I$, the decision tree that maximizes $\left|\sum_{i=1}^k c_i(G_q(A_i^I) - R(A_i^I))\right| $ is given by,
\begin{align}
    c_i=\begin{cases}
        +1, & G_q(A_i^I) \geq R(A_i^I)\\
        -1, & G_q(A_i^I) \leq R(A_i^I),
    \end{cases}
\end{align}
which results in,
\begin{align*}
\sup_{c\in\mathcal{C}}|E_{G_q}[c(X_g)] - E_R[c(X_r)]|
&=\max_{I\subset\{1,\dotsc,d\},|I|=k}2TV(\tilde{G_q}_I,\tilde{R}_I).
\end{align*}
\end{proof}

%% file: CVPR_2025/sec/X2_implementation_details.tex
\section{Implementation Details}
\label{appen:implementation_detail}

This section provides comprehensive implementation details to ensure the reproducibility of our experimental results. We begin by describing our methodology for measuring MPR, including data preprocessing, classifier training, and attribute detection. We then detail the implementation specifics of baseline methods and our fine-tuning approach, including hyperparameter settings and optimization choices.

\paragraph{Details on MPR measurement.}

We utilized the FairFace training dataset to train classifiers for detecting gender, age, and race attributes. For gender classification, we maintained the original binary categories (male and female) from the dataset. Race classification preserved the seven original categories: White, Black, Southeast Asian, Middle Eastern, East Asian, Latino Hispanic, and Indian. For age classification, since classifying fine-grained age groups is challenging, we simplified the task by binarizing the labels into ``young'' ($< 40$ years) and ``old'' ($\geq 40$ years) categories to improve classification reliability. The classifiers were trained in the CLIP embedding space using linear classifiers, and the optimal models were selected through a grid search algorithm. As a result, the classifiers achieved accuracies of 98\%, 95\%, and 77\% on the FairFace test dataset for gender, age, and race, respectively.



By default, we generated 1,000 images to measure MPR unless stated otherwise. We employed the Dlib face detector \cite{dlib09} to detect faces in the generated images and filtered to include only those images containing at least one detectable face. To estimate demographic attributes (gender, age, and race), we cropped the detected faces from the images. We applied our classifiers only to these cropped regions, which helps reduce noise from background elements. If multiple faces were detected in an image, we selected the largest face for analysis to ensure consistent evaluation. In contrast, for attribute-specific detection tasks (e.g., detecting the presence of a wheelchair or analyzing scene composition), we utilized the entire generated image without cropping to preserve all relevant visual information.

\paragraph{Implementation details of baselines and our finetuning method.} All baseline methods were implemented using their publicly released codebases to ensure fair comparison. For each baseline, we maintained their default hyperparameter configurations as specified in their respective papers and repositories. To ensure consistent evaluation criteria, we adapted all baseline methods to utilize FairFace statistics rather than assuming equal representation when applicable.
 

As described in Section \ref{sec:intervention_methods}, our fine-tuning approach incorporates the MPR term (Equation \ref{emp_mpr_def}) as a regularizer and leverages cached generations and functions $c$ to enable efficient gradient computation (detailed in Algorithm \ref{alg:ours}). The hyperparameter configuration is as follows: we used a learning rate of $0.0005$ with a warm-up phase and no subsequent scheduling, a mini-batch size of $8$ per iteration, and applied fine-tuning exclusively to the text encoder of the SD v1.4 model using LoRA for $10,000$ iterations. The regularization strength $\lambda$ was selected through a systematic search over $[0.1, 0.5, 1, 5, 10, 50, 100]$. To efficiently determine this value, we conducted preliminary training runs of $100$ iterations each and selected the largest value that demonstrated consistent MPR improvements compared to the pre-fine-tuned model. In our case, $\lambda = 0.5$ was chosen. The buffer sizes for storing generated images $B_{\text{MPR}}$ and functions $B_C$ were both set to $32$ to balance memory constraints with optimization stability.


\begin{algorithm}
    \footnotesize
        \caption{Finetuning algorithm for achieving MPR} \label{alg:ours}
        \SetKwInOut{Input}{Input}
        \SetKwInOut{Output}{Output}
        \Input { Model $\theta$, iteration $I$, mini-batch size $B$, MPR batch size $B_{\text{MPR}}$, $\widehat{C}$ size $B_C$, curation dataset $\mathcal{D}$}
        Set $\theta_0$ from a pre-trained diffusion model \\
        Set $\widehat{C }$ \& $\widehat{P} \leftarrow [\ ], [\ ]$\\
        \For{$t=0$ \KwTo $T-1$}
        {   
            \textcolor{orange}{\texttt{//\ Generate samples}} \;
            Generate $B$ images $X_B^t$, $X_B^o$ from $\theta_t$, $\theta_0$\;
            $\widehat{P}$.extend($X_B^t$) \;
            \If{$|\widehat{P}|>B_{\text{MPR}}$}{
                Pop some old images in $\widehat{P}$\;
            } 
            \textcolor{orange}{\texttt{//\ Calculate $c$}} \;
            $c_t \leftarrow \big| \frac{1}{k}\sum_{x_i\in \widehat{P}} c(x_i) - \frac{1}{m} \sum_{x_j \in \mathcal{D}} c(x_j)\big|$\;
            $\widehat{C}$.append($c_t$)\;
            \If{$|\widehat{C}|>B_C$}{
                Pop the oldest $c$ in $\widehat{C}$\;
            } 
            \textcolor{orange}{\texttt{//\ Gradient update $c$}} \;
            $\theta_{t+1} \leftarrow \theta_t - \alpha \nabla \mathcal{L}_{\text{obj ver1}}(\theta_t)$\;
        }        
        \Output {$\theta_T$}
\end{algorithm}


%% file: CVPR_2025/sec/X3_additional_results.tex
\section{Additional Experimental Results}
\label{appen:additional_experiments}

In this section, we present detailed experimental analyses that complement and expand upon the results shown in our main paper. We examine the performance characteristics of our MPR framework across multiple dimensions, including (i.) comprehensive trait representation analysis using additional state-of-the-art diffusion models (ii.) analysis of empirical vs. true MPR gaps across different function classes, (iii.) quantitative examination of contextual representation in historical settings, such as the ``computer programmer for ENIAC'' and (iv.) supplementary qualitative results. Each subsection provides experimental evidence supporting our theoretical framework while revealing new insights about representational biases in text-to-image systems.



\subsection{MPRs for Traits on Advanced Text-to-Image Models}

\input{CVPR_2025/tables/table5}

While our main paper examined trait biases across three baseline models (SD v1.4, SD v2.1, SDXL), here we extend this analysis to include four state-of-the-art models: LCM-SDXL \cite{luo2023latent}, Stable Cascade \cite{pernias2023wurstchen}, Playground v2.5 \cite{li2024playground}, and PixArt-$\Sigma$ \cite{chen2024pixart}. This broader analysis helps understand whether other recent architectural and training advances have addressed representational biases.

Table \ref{tab:mpr_trait_supple} presents MPR measurements using identical experimental settings as Table \ref{tab:mpr_trait} in the main paper. Our findings reveal several key insights:

\begin{itemize}
    \item Despite their documented improvements in image quality and generation capabilities, these newer models exhibit bias levels comparable to or sometimes exceeding those of SDXL across all six examined traits.
    \item We observe consistent bias patterns across all models regardless of their architectural differences. For example, decision trees consistently select ``white'' as a splitting criterion when evaluating ``attractive'' trait bias and ``black'' when evaluating ``thug'' trait bias. This consistency suggests these biases may stem from deeper societal stereotypes present in training data rather than specific architectural choices.
\end{itemize}

These findings emphasize that advances in model architecture and image generation quality do not automatically translate to improved representational fairness. They underscore the critical need for explicitly incorporating intersectional fairness considerations into model development rather than treating them as properties that will naturally emerge from technical improvements.

\subsection{Maximum Gap between Empirical and True MPRs Depending on the Function Class}
\label{appen:maximum_gap}
\begin{figure}
    \centering
    \includegraphics[width=0.5\linewidth]{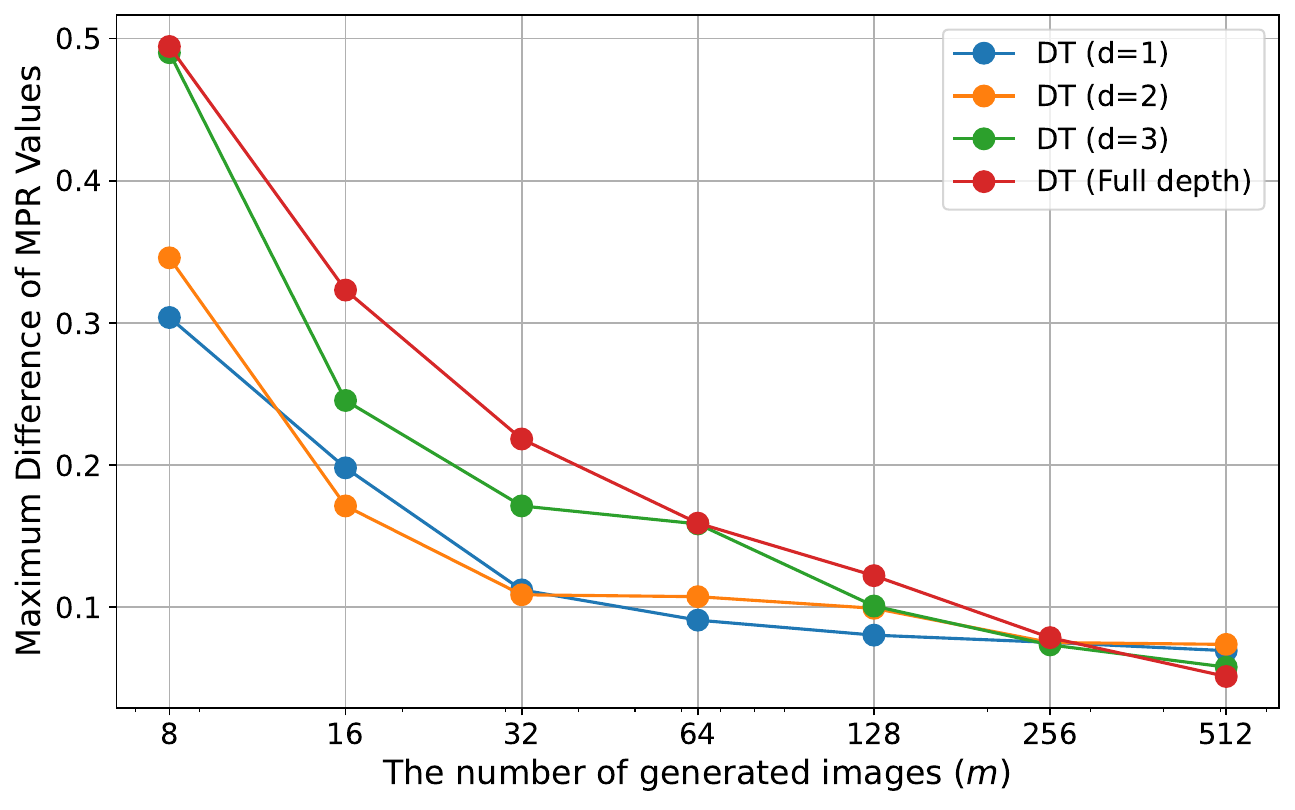}
    \caption{\small{\bf Maximum deviation between the estimated MPRs and the true MPR depending on decision tree depth and the number of generated images.} The number following ``DT'' indicates the depth of the decision tree, and the maximum deviation is calculated after 30 repetitions of the estimation process.}
    \label{fig:max_gap}
\end{figure}

In Proposition~\ref{gen}, we established the relationship between the number of images of $\widehat{G}_q$ and $\widehat{R}$, the complexity of the function class $\mathcal{C}$, and the approximation error of $\text{MPR}(\mathcal{C}, \widehat{G}_q, \widehat{R})$. Specifically, we demonstrated that, in the worst case, there is a trade-off between function complexity and the number of images to achieve the same level of accuracy in MPR estimation.

To support this experimentally, we analyzed how the accuracy of MPR estimation changes with respect to function complexity and the number of images when estimating the MPR value for the CEO using the SD v1.4 model. To calculate the approximation error for each trial of estimating an MPR value, \ie, the LHS in eq. \ref{eq:prop1}, we assume that the MPR value calculated using 5,000 generated images is an accurate approximation of the true MPR. For the given number of generated images --$m$-- we generated images, estimated the MPR, and repeated this process 30 times. We then reported the maximum deviation from the assumed true MPR. In this experiment, we used the FairFace test dataset as the reference data and calculated MPR using gender, age, and race group labels with decision tree functions. 

As shown in Figure \ref{fig:max_gap}, the deviation from the true MPR consistently decreases as the number of images increases. Furthermore, we observe that the deviation becomes larger when using deeper decision trees, \ie, more complex functions. These findings directly support the arguments in Proposition \ref{gen}, implying that while MPR can capture more nuanced biases with higher function complexity, the cost of obtaining sufficient images to maintain accuracy also increases.





\subsection{MPR Results of Other Baseline Methods for ``Computer programmer for ENIAC''}\label{append:eniac}

\input{CVPR_2025/tables/table4}

Building upon the results presented in Table \ref{tab:contextual_query} of the main paper, we evaluated the performance of additional baseline methods in representing ENIAC programmers. Table \ref{tab:contextual_query_supple} presents MPR values measured against both equal gender distribution and historically accurate (Google Images) reference sets. Our expanded analysis reveals:

\begin{itemize}
    \item All baseline methods successfully reduce bias when evaluated against equal gender representation, with FairDiffusion and Entigen achieving particularly low MPR scores (0.03 and 0.04, respectively).
    \item However, when evaluated against the historically accurate reference distribution (which consists solely of female programmers), all methods show substantial representational disparities. Even the best-performing methods (FairDiffusion and Entigen) exhibit MPR values above 0.30. One reason for the increase in MPR with non-uniform reference statistics in existing diffusion models (even with fairness interventions) is that they are trained to ensure equal representation and not contextual representation. 
    \item Notably, ITI-GEN, which performs well in other contexts, shows the highest disparity (0.83) against historical accuracy, suggesting that methods optimized for general fairness might inadvertently work against accurate historical representation.
\end{itemize}

These results highlight a fundamental tension between different fairness objectives: while these methods effectively promote gender equality in general, they may inadvertently diminish historically significant representations of women in computing. This underscores the importance of context-aware evaluation metrics and the need for more nuanced approaches to bias mitigation.




\subsection{Qualitative Results}

In this section, we show that an MPR-optimized model demonstrates significant improvements in achieving balanced representation across intersectional groups. Specifically, we fine-tuned Stable Diffusion v1.4 using our MPR-based optimization approach to evaluate this capability. As shown in Figures \ref{fig:occupation_images} and, when generating images for the prompt ``a portrait photo of a \{pilot, chef, flight attendant, housekeeper, taxi driver, nurse, therapist\}'' our fine-tuned model produces diverse, high-quality images that span multiple demographic dimensions, including gender, age, and race. While the original SD v1.4 predominantly generates images of white male pilots, and FairDiffusion primarily addresses gender balance, our MPR-optimized version effectively captures broader intersectional diversity. The qualitative results in Figure \ref{fig:occupation_images} and \ref{fig:occupation_images2} (split into two figures for formatting purposes) validate our quantitative findings, showing that MPR optimization can successfully balance representation across demographic groups while preserving image quality and maintaining appropriate professional context.
\begin{figure}
    \centering
    \vspace{.1cm}
    
    \includegraphics[width=0.6\linewidth]{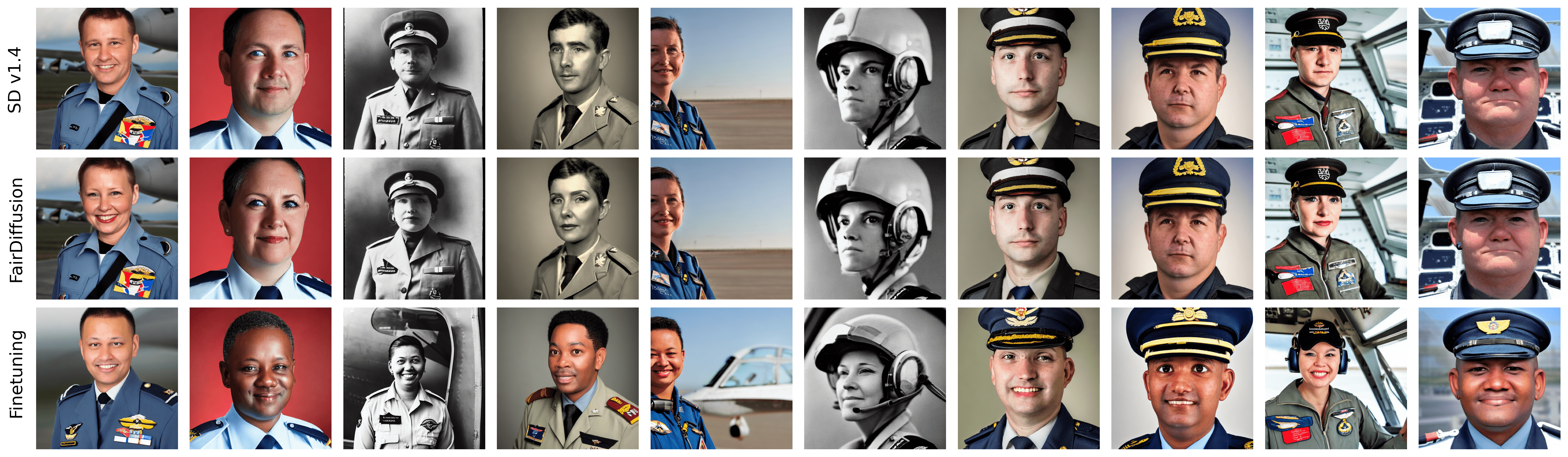}
    
    \vspace{.1cm}
    
    \includegraphics[width=0.6\linewidth]{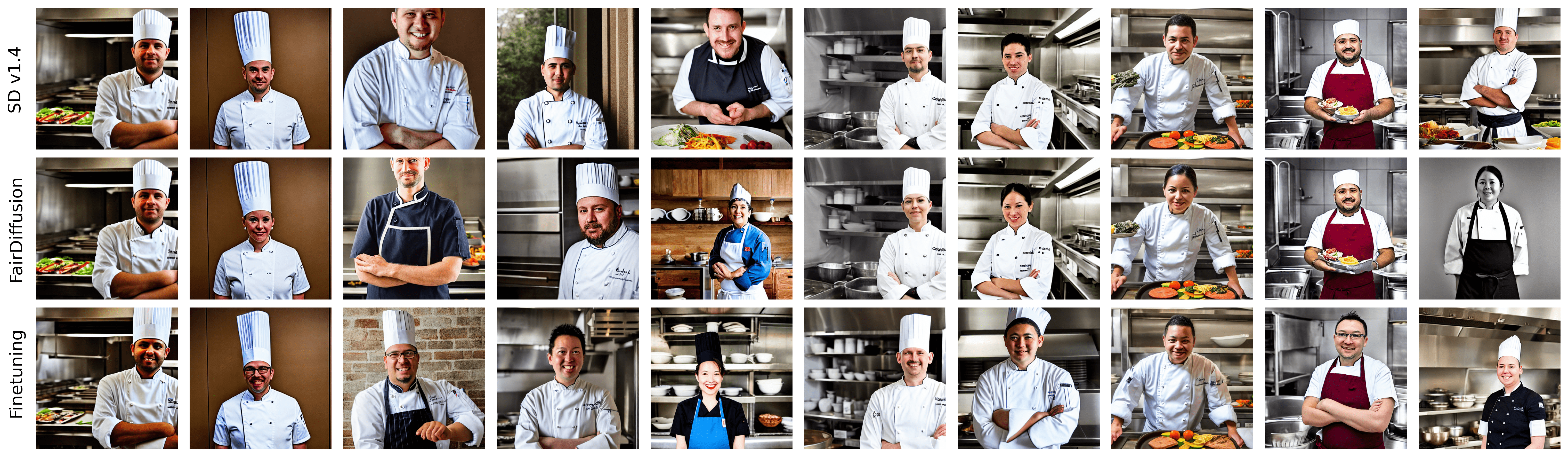}
    
    \vspace{.1cm}
    
    \includegraphics[width=0.6\linewidth]{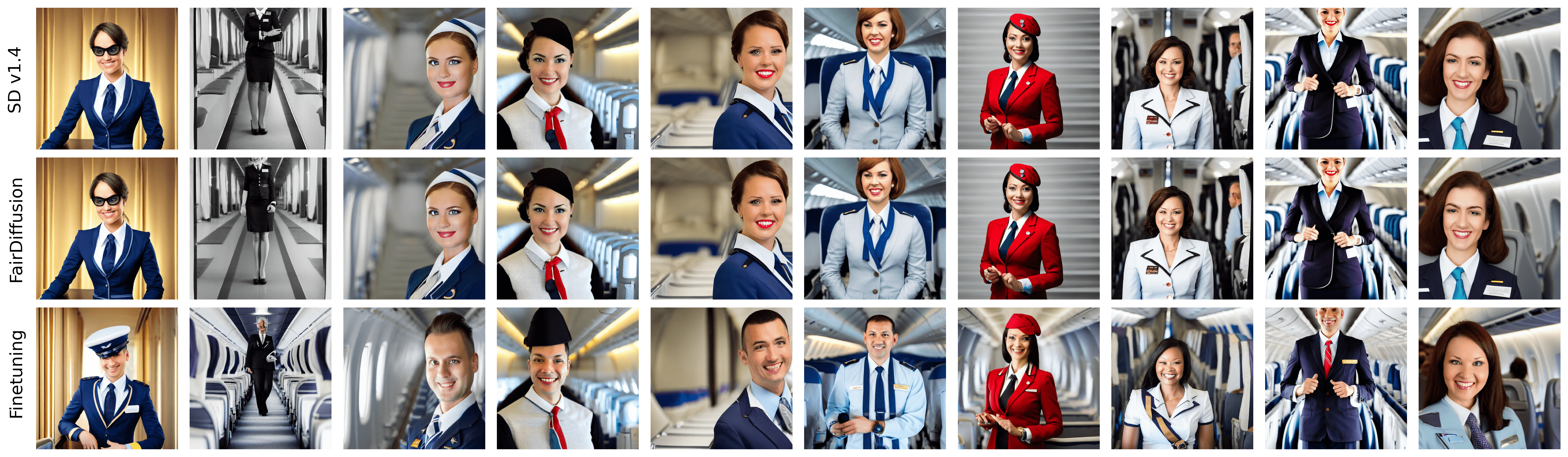}

    \caption{\small{Qualitative comparison of images generated for the prompt \emph{a portrait photo of a \{pilot, chef, flight attendant, housekeeper, taxi driver, nurse, therapist\}} across different methods. (Top row) Images generated by the original SD v1.4 model show a strong bias toward white male pilots, lacking demographic diversity. (Middle row) Images generated by FairDiffusion achieve better gender balance but still show limited representation across intersectional groups. (Bottom row) Images generated by SD v1.4 fine-tuned with MPR.}}
    \label{fig:occupation_images}
\end{figure}

\begin{figure}
    \centering

    \vspace{.1cm}
    
    \includegraphics[width=0.6\linewidth]{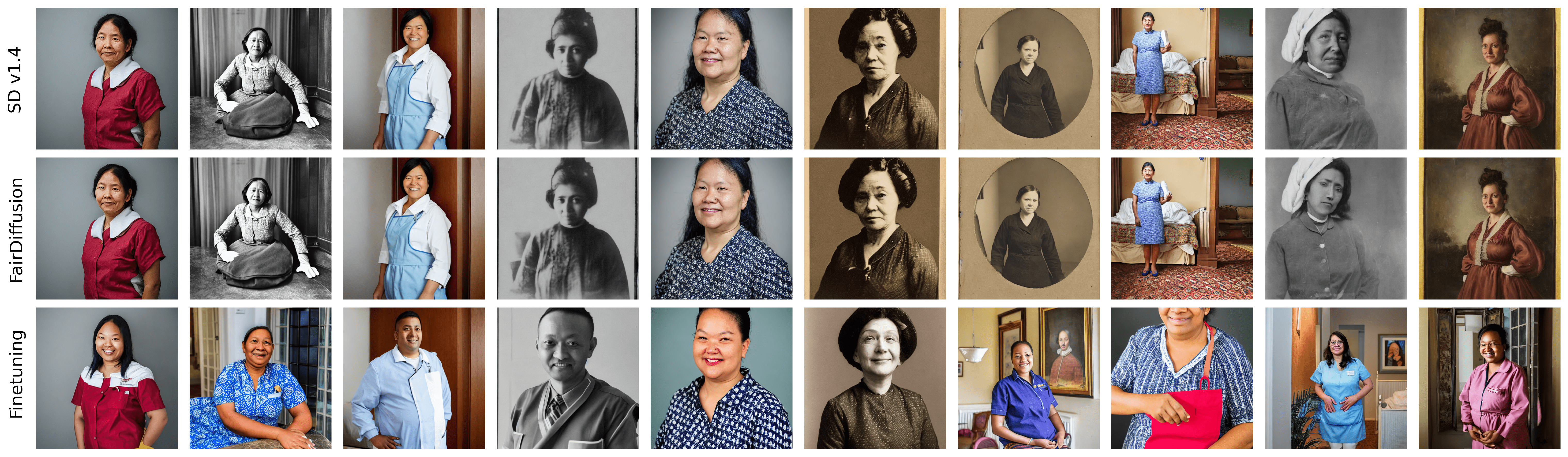}
    \vspace{.1cm}
    
    \includegraphics[width=0.6\linewidth]{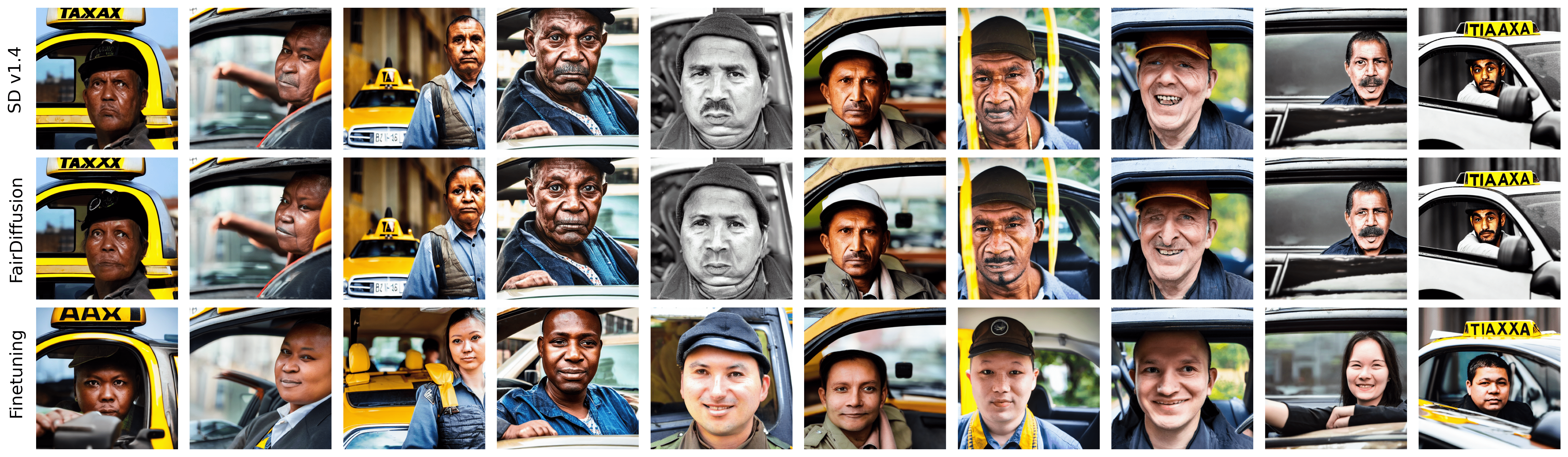}
    
    \vspace{.1cm}
    
    \includegraphics[width=0.6\linewidth]{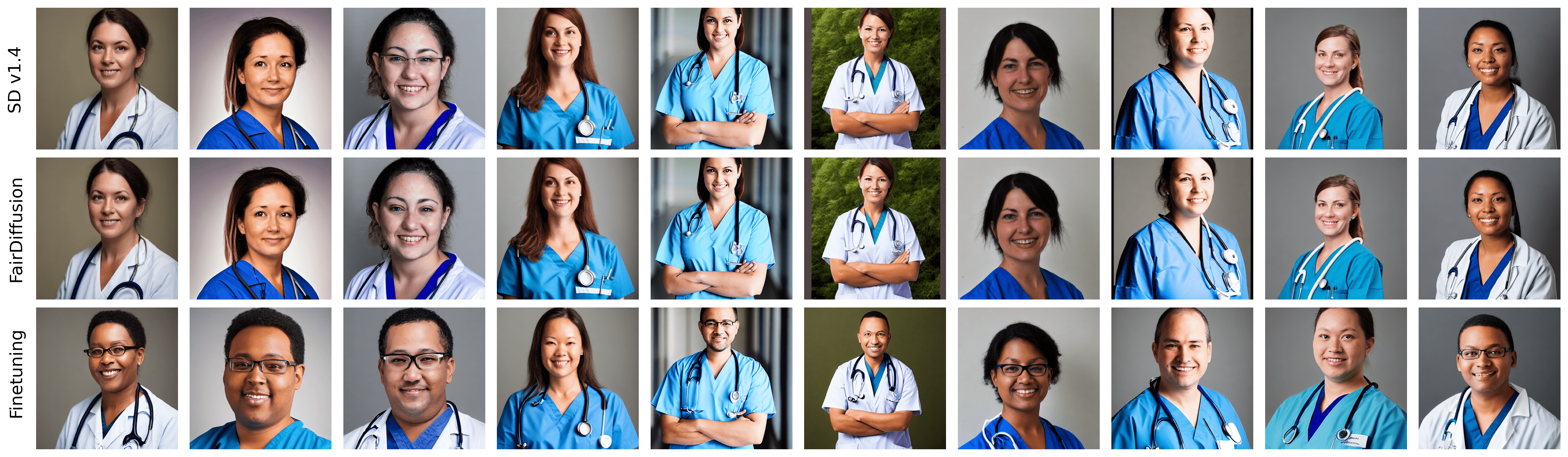}
    \vspace{.1cm}
    
    \includegraphics[width=0.6\linewidth]{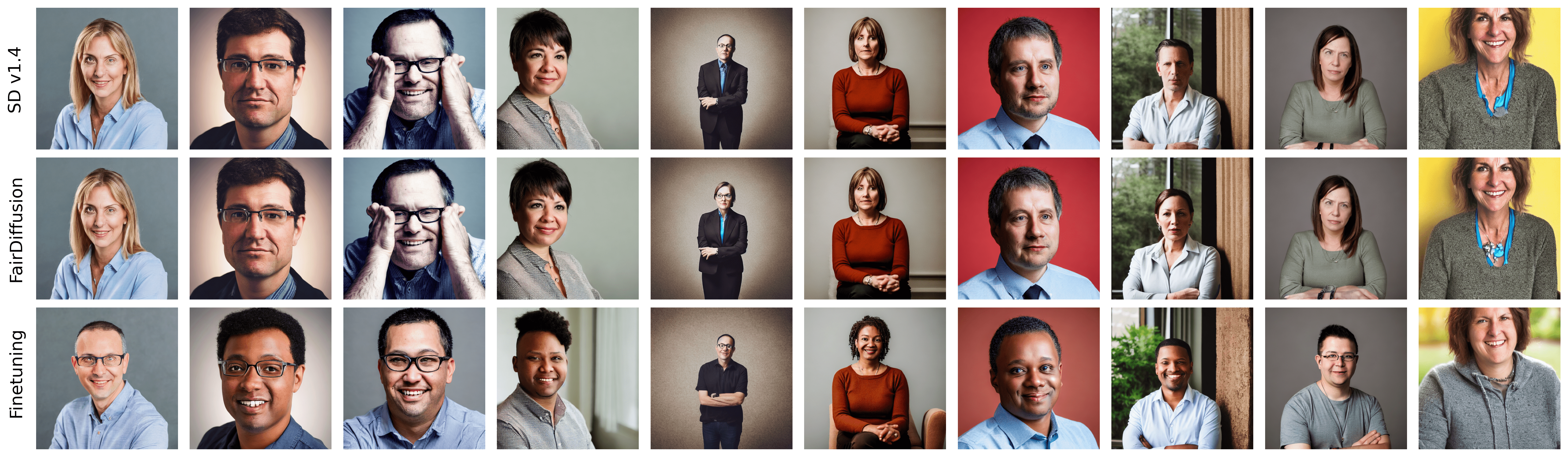}
    
    \caption{\small{Continuation of Figure \ref{fig:occupation_images}. Qualitative comparison of images generated for the prompt \emph{a portrait photo of a \{housekeeper, taxi driver, nurse, therapist\}} across different methods.}}
    \label{fig:occupation_images2}
\end{figure}

\subsection{Sensitivity Analysis on the Regularization Strength in Our Finetuning Method} \label{subsec:lambda analysis}
We investigate how the regularization strength $\lambda$ in our objective affects the performance of the model, in terms of representational fairness and image quality. Table \ref{tab:lambda_analysis} presents the MPR values and CLIP scores for different values of $lambda$. As $\lambda$ increases, we observe a consistent decrease in MPR and an increase in CLIP. This trend demonstrates that the regularization term effectively guides the model toward generating more representative and semantically coherent outputs. 
\begin{table}[t!]
    \caption{MPR and CLIP scores by varying $\lambda$ of our finetuning method}
    \centering
    \begin{tabular}{ccccc|c}
    \toprule
        $\lambda$ &   $\left(\tfrac{1}{2}\right)^{\frac{3}{2}}$ & $\left(\tfrac{1}{2}\right)^1$ & $\left(\tfrac{1}{2}\right)^{\frac{1}{2}}$ & $\left(\tfrac{1}{2}\right)^0$ & SD v1.4 \\\midrule
         MPR (GAR) ($\downarrow$) & 0.235 & 0.252 & 0.267 & 0.335 & 1.89 \\
         CLIP score ($\uparrow$)& 0.292 & 0.300 & 0.303 & 0.306 & 0.311 \\
         \bottomrule
    \end{tabular}
    \label{tab:lambda_analysis}
\end{table}

%% file: CVPR_2025/tables/table5.tex
\begin{table*}[t!]
\scriptsize
\setlength{\tabcolsep}{2pt}
\centering
\caption{\textbf{MPR results for several traits}. Every MPR is obtained with groups of ``Male'', ``Old'' and ``race'' and decision trees. The splits highlighted in bold font indicate those used in the decision trees with a depth of 1. The number in parenthesis represents the standard deviation by bootstrapping.}
\begin{tabular}{@{}llcccccc@{}}
\toprule
                                                         &                                    & Attractive                                                                       & Emotional                   & Exotic                                                                        & Poor                                                                                   & Terrorist                                                                              & Thug                                                                       \\ \midrule \midrule
\multirow{2}{*}{LCM-SDXL}                                 & DT$(d=1)$                                & 0.60 ($\pm$0.01)
                                                                  & 0.44 ($\pm$0.01)
             & 0.51 ($\pm$0.02)                                                               & 
0.44 ($\pm$0.01)                                                                  & 
0.44 ($\pm$0.01)                                                      & 
0.58 ($\pm$0.02) \\
                                                         & DT$(d=3)$                                & 0.72 ($\pm$0.01)                                                                  & 0.63 ($\pm$0.01) & 0.64 ($\pm$0.01) & 0.66 ($\pm$0.01) & 0.62 ($\pm$0.01) & 0.69 ($\pm$0.01)\\\midrule 
\multirow{2}{*}{Stable Cascade}                                   & DT$(d=1)$                                & 0.57 ($\pm$0.01)                                                                 & 0.55 ($\pm$0.01) & 0.46 ($\pm$0.01) & 0.54 ($\pm$0.01) & 0.83 ($\pm$0.00) & 0.41($\pm$0.01) \\ 
& DT$(d=3)$ & 0.81 ($\pm$0.00) & 0.63 ($\pm$0.01) & 0.64 ($\pm$0.01) & 0.59 ($\pm$0.01) & 0.87 ($\pm$0.00) & 0.60 ($\pm$0.01)\\\midrule
\multirow{2}{*}{Playground v2.5}                                    & DT$(d=1)$                                & 0.49 ($\pm$0.01)                                                                  & 0.41 ($\pm$0.01) & 0.79 ($\pm$0.01) & 0.74 ($\pm$0.01) & 0.47 ($\pm$0.00) & 0.66 ($\pm$0.01) \\
                                                         & DT$(d=3)$  & 0.73 ($\pm$0.00) & 0.55 ($\pm$0.01) & 0.84 ($\pm$0.01) & 0.80 ($\pm$0.01) & 0.65 ($\pm$0.01) & 0.85 ($\pm$0.00) \\\midrule 
\multirow{2}{*}{PixArt-$\Sigma$}                                   & DT$(d=1)$                                & 0.58 ($\pm$0.01)                                                                  & 0.55 ($\pm$0.01) & 0.46 ($\pm$0.01) & 0.54 ($\pm$0.01) & 0.83 ($\pm$0.00) & 0.41 ($\pm$0.01)\\ 
& DT$(d=3)$                                & 0.71 ($\pm$0.01)                                                                  & 0.63 ($\pm$0.01) & 0.64 ($\pm$0.01) & 0.59 ($\pm$0.01) & 0.87 ($\pm$0.00) & 0.60 ($\pm$0.01)\\\midrule\midrule
\multicolumn{2}{c}{\multirow{3}{*}{\begin{tabular}[c]{@{}c@{}}Splits of DT$(d=3)$ \\ LCM-SDXL\end{tabular}}} & \multirow{3}{*}{\begin{tabular}[c]{@{}c@{}}Male\\ Latino  \\ \textbf{White}\end{tabular}} & \multirow{3}{*}{\begin{tabular}[c]{@{}c@{}} Old\\ \textbf{White} \\ Latino\end{tabular}}& \multirow{3}{*}{\begin{tabular}[c]{@{}c@{}}Old\\ Latino \\ \textbf{Indian}\end{tabular}} & \multirow{3}{*}{\begin{tabular}[c]{@{}c@{}}\textbf{Old}\\ Black \\ Indian\end{tabular}} & \multirow{3}{*}{\begin{tabular}[c]{@{}c@{}}\textbf{Male}\\ Old \\ East Asian\end{tabular}} & \multirow{3}{*}{\begin{tabular}[c]{@{}c@{}}Male\\ Latino \\ \textbf{Black}\end{tabular}} \\ 
\multicolumn{2}{l}{}                                                                          &                                                                                  &                             &                                                                               &                                                                                        &                                                                                        &                                                                            \\ 
\multicolumn{2}{l}{}                                                                          &                                                                                  &                             &                                                                               &                                                                                        &                                                                                        &                                                                            \\ \midrule
\multicolumn{2}{c}{\multirow{3}{*}{\begin{tabular}[c]{@{}c@{}}Splits of DT$(d=3)$ \\ Stable Cascade\end{tabular}}} & \multirow{3}{*}{\begin{tabular}[c]{@{}c@{}}Male\\ Latino\\ \textbf{White}\end{tabular}} &\multirow{3}{*}{\begin{tabular}[c]{@{}c@{}}Old\\ Latino \\ \textbf{White}\end{tabular}} & \multirow{3}{*}{\begin{tabular}[c]{@{}c@{}}\textbf{Male}\\ Indian \\ Latino\end{tabular}} & \multirow{3}{*}{\begin{tabular}[c]{@{}c@{}}\textbf{Old}\\ Middle Eastern \\ Southeast Asian\end{tabular}} & \multirow{3}{*}{\begin{tabular}[c]{@{}c@{}}Male\\ Old \\ \textbf{Indian}\end{tabular}} & \multirow{3}{*}{\begin{tabular}[c]{@{}c@{}}\textbf{Male}\\ Old\\ East Asian\end{tabular}} \\
\multicolumn{2}{l}{}                                                                          &                                                                                  &                             &                                                                               &                                                                                        &                                                                                        &                                                                            \\ 
\multicolumn{2}{l}{}                                                                          &                                                                                  &                             &                                                                               &                                                                                        &                                                                                        &                                                                            \\ \midrule
\multicolumn{2}{c}{\multirow{3}{*}{\begin{tabular}[c]{@{}c@{}}Splits of DT$(d=3)$ \\ Playground v2.5\end{tabular}}} & \multirow{3}{*}{\begin{tabular}[c]{@{}c@{}}Old\\ Latino \\ \textbf{White}\end{tabular}} & \multirow{3}{*}{\begin{tabular}[c]{@{}c@{}} \textbf{Male} \\ Middle Eastern\\ East Asian \end{tabular}} & \multirow{3}{*}{\begin{tabular}[c]{@{}c@{}}Male\\ Old \\ \textbf{Indian}\end{tabular}} & \multirow{3}{*}{\begin{tabular}[c]{@{}c@{}}Male\\ \textbf{Indian} \\ - \end{tabular}} & \multirow{3}{*}{\begin{tabular}[c]{@{}c@{}}\textbf{Male}\\ Indian \\ White\end{tabular}} & \multirow{3}{*}{\begin{tabular}[c]{@{}c@{}}Male\\ Indian \\ \textbf{Black}\end{tabular}} \\
\multicolumn{2}{l}{}                                                                          &                                                                                  &                             &                                                                               &                                                                                        &                                                                                        &                                                                            \\
\multicolumn{2}{l}{}                                                                          &                                                                                  &                             &                                                                               &                                                                                        &                                                                                        &                                                                            \\  \midrule
\multicolumn{2}{c}{\multirow{3}{*}{\begin{tabular}[c]{@{}c@{}}Splits of DT$(d=3)$ \\ PixArt-$\Sigma$\end{tabular}}} & \multirow{3}{*}{\begin{tabular}[c]{@{}c@{}}Male\\ Old \\ \textbf{White}\end{tabular}} & \multirow{3}{*}{\begin{tabular}[c]{@{}c@{}} Male \\ Latino\\ \textbf{White} \end{tabular}} & \multirow{3}{*}{\begin{tabular}[c]{@{}c@{}}Old \\ \textbf{Indian} \\Black \end{tabular}} & \multirow{3}{*}{\begin{tabular}[c]{@{}c@{}}\textbf{Old}\\ Indian \\ Black \end{tabular}} & \multirow{3}{*}{\begin{tabular}[c]{@{}c@{}}\textbf{Male}\\ Indian \\ Middle Eastern\end{tabular}} & \multirow{3}{*}{\begin{tabular}[c]{@{}c@{}}Male\\ \textbf{White} \\ Black\end{tabular}} \\
\multicolumn{2}{l}{}                                                                          &                                                                                  &                             &                                                                               &                                                                                        &                                                                                        &                                                                            \\
\multicolumn{2}{l}{}                                                                          &                                                                                  &                             &                                                                               &                                                                                        &                                                                                        &                                                                            \\  \bottomrule
\end{tabular}
\label{tab:mpr_trait_supple}
\end{table*}

%% file: CVPR_2025/tables/table4.tex
\begin{table}
\centering
\small
\caption{\textbf{MPR values of different diffusion models on a balanced (left) and contextual (right) curation set.}}
\label{tab:contextual_query_supple}
\begin{tabular}{@{}lll@{}}
\toprule
Reference     & Equal          & Google Images  \\ \midrule \midrule
SD v1.4       & 0.40 {\scriptsize($\pm$ 0.01)} & 0.69 {\scriptsize($\pm$ 0.01)} \\ 
FairDiffusion \cite{friedrich2023fair} & 0.03 {\scriptsize($\pm$ 0.01)} & 0.32 {\scriptsize($\pm$ 0.01)} \\
Entigen \cite{bansal2022well} & 0.04 {\scriptsize($\pm$ 0.02)} & 0.33 {\scriptsize($\pm$ 0.01)} \\
UCE \cite{gandikota2024unified} & 0.19 {\scriptsize($\pm$ 0.01)} & 0.48 {\scriptsize($\pm$ 0.01)} \\
ITI-GEN \cite{zhang2023iti}           & 0.33 {\scriptsize($\pm$ 0.01)} & 0.83 {\scriptsize($\pm$ 0.01)}  \\
\bottomrule
\end{tabular}
\end{table}


%% file: CVPR_2025/sec/X4_sample_sizes.tex
\begin{figure}[t!]
    \centering \includegraphics[width=0.6\linewidth]{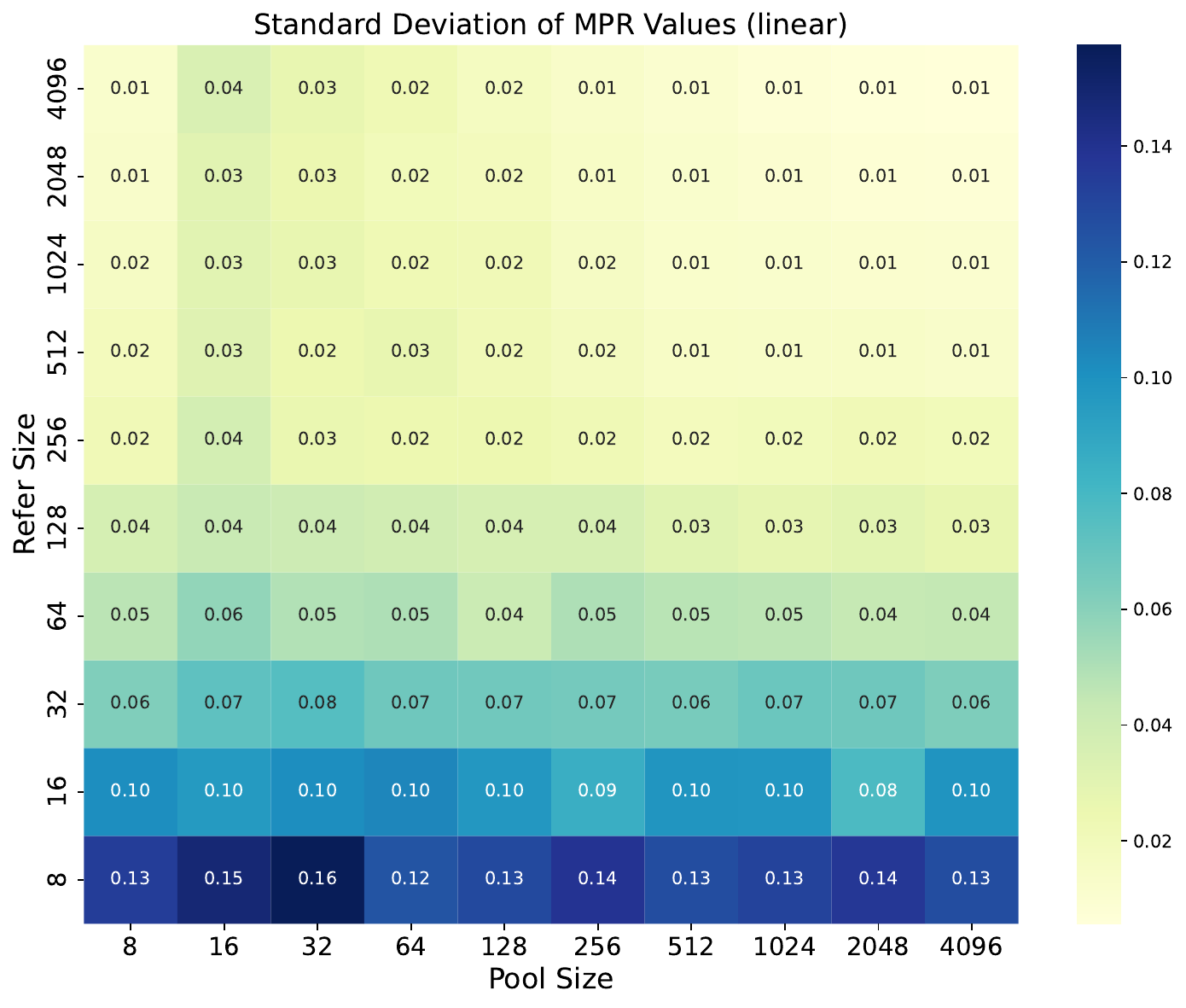}
    \caption{\small \textbf{Heatmap of MPR standard deviations obtained by bootstrapping.} Each value represents the MPR calculated using a linear classifier function class, considering gender, age, and race as group attributes.}
    \label{fig:std_heatmap}
\end{figure}

\section{Discussion on the practical selection of sample sizes in the MPR framework}
\label{appen:practical_guidelines}

We identify two primary real-world scenarios where the MPR framework can be utilized to evaluate text-to-image models. In the first scenario, we assess whether a generative model $f$ sampling from distribution $G$ meets a specified fairness threshold $\rho$ (e.g., a representation target mandated by policymakers) by testing if $\mathrm{MPR}(\mathcal{C}, G, R) \leq \rho$. In the second scenario, we compare two different generative models $f$ and $g$, sampling from distributions $G$ and $G'$ respectively, by evaluating whether $\mathrm{MPR}(\mathcal{C}, G, R) \leq \mathrm{MPR}(\mathcal{C}, G', R)$ and vice versa. For the threshold comparison scenario, we employ a one-sided t-test to determine if the model achieves representation below $\rho$. For model comparison, we utilize a two-sided t-test to assess relative performance between models. In both cases, when the estimated MPR value approaches $\rho$ or when two models yield similar MPR values, higher precision estimates become crucial, necessitating lower standard deviation in our measurements. 

As established in Proposition \ref{gen}, we can achieve this increased precision by expanding the number of sampled images. This allows MPR users to control the trade-off between computational cost and estimation accuracy by adjusting the number of generated and reference images based on their specific requirements. However, since the true distributions $G$ and $R$ are unknown in practice, we cannot directly calculate the standard deviation. Instead, we employ bootstrapping to estimate these values empirically. Figure \ref{fig:std_heatmap} visualizes the relationship between sample sizes and estimation precision, showing standard deviation values estimated through bootstrapping as a function of the number of generated --$k$-- and reference --$m$-- images. The observed decrease in standard deviation with increasing $k$ and $m$ aligns with our theoretical predictions from Proposition \ref{gen}, providing practical guidance for sample size selection in MPR evaluation.

%% file: CVPR_2025/main.bbl
\begin{thebibliography}{67}
\providecommand{\natexlab}[1]{#1}
\providecommand{\url}[1]{\texttt{#1}}
\expandafter\ifx\csname urlstyle\endcsname\relax
  \providecommand{\doi}[1]{doi: #1}\else
  \providecommand{\doi}{doi: \begingroup \urlstyle{rm}\Url}\fi

\bibitem[UN_()]{UN_populationdivision}
United nations department of economic and social affairs: Population division.
\newblock \url{https://population.un.org/wpp/}.
\newblock Accessed: 2024-11-14.

\bibitem[cdc()]{cdc}
Centers for disease control and prevention.
\newblock \url{https://dhds.cdc.gov/SP}.
\newblock Accessed Sep 11, 2024.

\bibitem[eur()]{eurostats_data}
Eurostats data.
\newblock \url{https://ec.europa.eu/eurostat/web/population-demography/demography-population-stock-balance/database}.
\newblock Accessed: 2024-11-14.

\bibitem[ilo()]{ilostat_data}
International labour organization statistics.
\newblock \url{https://ilostat.ilo.org/}.
\newblock Accessed: 2024-11-14.

\bibitem[oec()]{oecd_data}
Oecd data.
\newblock \url{https://www.oecd.org/en/data.html}.
\newblock Accessed: 2024-11-14.

\bibitem[us_()]{us_bureau}
Us bureau of labor statistics. employed persons by detailed occupation, sex, race, and hispanic or latino ethnicity.
\newblock \url{https://www.bls.gov/cps/cpsaat11.htm}.
\newblock Accessed: 2024-11-09.

\bibitem[wor()]{worldbank_data}
World bank open data.
\newblock \url{https://data.worldbank.org/}.
\newblock Accessed: 2024-11-14.

\bibitem[{Adobe}(2024)]{adobe_firefly}
{Adobe}.
\newblock Firefly, 2024.
\newblock Accessed Sep 7, 2024. \url{https://www.adobe.com/products/firefly/features/text-to-image.html}.

\bibitem[Bansal et~al.(2022)Bansal, Yin, Monajatipoor, and Chang]{bansal2022well}
Hritik Bansal, Da Yin, Masoud Monajatipoor, and Kai-Wei Chang.
\newblock How well can text-to-image generative models understand ethical natural language interventions?
\newblock \emph{arXiv preprint arXiv:2210.15230}, 2022.

\bibitem[Bartik(2013)]{bartik2013pioneer}
Jean~Jennings Bartik.
\newblock \emph{Pioneer programmer: Jean Jennings Bartik and the computer that changed the world}.
\newblock Truman State University Press, 2013.

\bibitem[Bianchi et~al.(2023)Bianchi, Kalluri, Durmus, Ladhak, Cheng, Nozza, Hashimoto, Jurafsky, Zou, and Caliskan]{bianchi2023easily}
Federico Bianchi, Pratyusha Kalluri, Esin Durmus, Faisal Ladhak, Myra Cheng, Debora Nozza, Tatsunori Hashimoto, Dan Jurafsky, James Zou, and Aylin Caliskan.
\newblock Easily accessible text-to-image generation amplifies demographic stereotypes at large scale.
\newblock In \emph{ACM Conference on Fairness, Accountability, and Transparency}, pages 1493--1504, 2023.

\bibitem[Borgwardt et~al.(2006)Borgwardt, Gretton, Rasch, Kriegel, Sch{\"o}lkopf, and Smola]{mmd1}
Karsten~M Borgwardt, Arthur Gretton, Malte~J Rasch, Hans-Peter Kriegel, Bernhard Sch{\"o}lkopf, and Alex~J Smola.
\newblock Integrating structured biological data by kernel maximum mean discrepancy.
\newblock \emph{Bioinformatics}, 22\penalty0 (14):\penalty0 e49--e57, 2006.

\bibitem[Boucheron et~al.(2013)Boucheron, Lugosi, and Massart]{boucheron2013concentration}
St{\'e}phane Boucheron, G{\'a}bor Lugosi, and Pascal Massart.
\newblock Concentration inequalities: A nonasymptotic theory of independence.
\newblock 2013.

\bibitem[Chen et~al.(2024)Chen, Ge, Xie, Wu, Yao, Ren, Wang, Luo, Lu, and Li]{chen2024pixart}
Junsong Chen, Chongjian Ge, Enze Xie, Yue Wu, Lewei Yao, Xiaozhe Ren, Zhongdao Wang, Ping Luo, Huchuan Lu, and Zhenguo Li.
\newblock Pixart-$\sigma$: Weak-to-strong training of diffusion transformer for 4k text-to-image generation.
\newblock \emph{arXiv preprint arXiv:2403.04692}, 2024.

\bibitem[Chinchure et~al.(2024)Chinchure, Shukla, Bhatt, Salij, Hosanagar, Sigal, and Turk]{chinchure2024tibet}
Aditya Chinchure, Pushkar Shukla, Gaurav Bhatt, Kiri Salij, Kartik Hosanagar, Leonid Sigal, and Matthew Turk.
\newblock Tibet: Identifying and evaluating biases in text-to-image generative models.
\newblock In \emph{European Conference on Computer Vision}, pages 429--446. Springer, 2024.

\bibitem[Cho et~al.(2023)Cho, Zala, and Bansal]{cho2023dall}
Jaemin Cho, Abhay Zala, and Mohit Bansal.
\newblock Dall-eval: Probing the reasoning skills and social biases of text-to-image generation models.
\newblock In \emph{European Conference on Computer Vision}, pages 3043--3054, 2023.

\bibitem[Choi et~al.(2024)Choi, Park, Kim, Lee, and Park]{choi2024fair}
Yujin Choi, Jinseong Park, Hoki Kim, Jaewook Lee, and Saerom Park.
\newblock Fair sampling in diffusion models through switching mechanism.
\newblock In \emph{AAAI Conference on Artificial Intelligence}, pages 21995--22003, 2024.

\bibitem[Chuang et~al.(2023)Chuang, Jampani, Li, Torralba, and Jegelka]{chuang2023debiasing}
Ching-Yao Chuang, Varun Jampani, Yuanzhen Li, Antonio Torralba, and Stefanie Jegelka.
\newblock Debiasing vision-language models via biased prompts.
\newblock \emph{arXiv preprint arXiv:2302.00070}, 2023.

\bibitem[D'Inc{\`a} et~al.(2024{\natexlab{a}})D'Inc{\`a}, Peruzzo, Mancini, Xu, Goel, Xu, Wang, Shi, and Sebe]{d2024openbias}
Moreno D'Inc{\`a}, Elia Peruzzo, Massimiliano Mancini, Dejia Xu, Vidit Goel, Xingqian Xu, Zhangyang Wang, Humphrey Shi, and Nicu Sebe.
\newblock Openbias: Open-set bias detection in text-to-image generative models.
\newblock In \emph{IEEE/CVF Conference on Computer Vision and Pattern Recognition}, pages 12225--12235, 2024{\natexlab{a}}.

\bibitem[D'Inc{\`a} et~al.(2024{\natexlab{b}})D'Inc{\`a}, Peruzzo, Mancini, Xu, Goel, Xu, Wang, Shi, and Sebe]{dinca2024openbias}
Moreno D'Inc{\`a}, Elia Peruzzo, Massimiliano Mancini, Dejia Xu, Vidit Goel, Xingqian Xu, Zhangyang Wang, Humphrey Shi, and Nicu Sebe.
\newblock Openbias: Open-set bias detection in text-to-image generative models.
\newblock In \emph{IEEE/CVF Conference on Computer Vision and Pattern Recognition}, pages 12225--12235, 2024{\natexlab{b}}.

\bibitem[Fraser et~al.(2023{\natexlab{a}})Fraser, Kiritchenko, and Nejadgholi]{fraser2023diversity}
Kathleen~C Fraser, Svetlana Kiritchenko, and Isar Nejadgholi.
\newblock Diversity is not a one-way street: Pilot study on ethical interventions for racial bias in text-to-image systems.
\newblock \emph{IEEE/CVF International Conference on Computer Vision}, 2023{\natexlab{a}}.

\bibitem[Fraser et~al.(2023{\natexlab{b}})Fraser, Nejadgholi, and Kiritchenko]{fraser23friendly}
Kathleen~C Fraser, Isar Nejadgholi, and Svetlana Kiritchenko.
\newblock A friendly face: Do text-to-image systems rely on stereotypes when the input is under-specified?
\newblock In \emph{The AAAI-23 Workshop on Creative AI Across Modalities}, 2023{\natexlab{b}}.

\bibitem[Friedrich et~al.(2023)Friedrich, Brack, Struppek, Hintersdorf, Schramowski, Luccioni, and Kersting]{friedrich2023fair}
Felix Friedrich, Manuel Brack, Lukas Struppek, Dominik Hintersdorf, Patrick Schramowski, Sasha Luccioni, and Kristian Kersting.
\newblock Fair diffusion: Instructing text-to-image generation models on fairness.
\newblock \emph{arXiv preprint arXiv:2302.10893}, 2023.

\bibitem[Gandikota et~al.(2024)Gandikota, Orgad, Belinkov, Materzy{\'n}ska, and Bau]{gandikota2024unified}
Rohit Gandikota, Hadas Orgad, Yonatan Belinkov, Joanna Materzy{\'n}ska, and David Bau.
\newblock Unified concept editing in diffusion models.
\newblock In \emph{IEEE/CVF Winter Conference on Applications of Computer Vision}, pages 5111--5120, 2024.

\bibitem[{Google}(2023)]{google_gemini}
{Google}.
\newblock Introducing gemini: our largest and most capable ai model, 2023.
\newblock Accessed Sep 7, 2024. \url{https://blog.google/technology/ai/google-gemini-ai/}.

\bibitem[Gretton et~al.(2012)Gretton, Borgwardt, Rasch, Sch{\"o}lkopf, and Smola]{mmd2}
Arthur Gretton, Karsten~M Borgwardt, Malte~J Rasch, Bernhard Sch{\"o}lkopf, and Alexander Smola.
\newblock A kernel two-sample test.
\newblock \emph{The Journal of Machine Learning Research}, 13\penalty0 (1):\penalty0 723--773, 2012.

\bibitem[Hamidi et~al.(2018)Hamidi, Scheuerman, and Branham]{hamidi2018gender}
Foad Hamidi, Morgan~Klaus Scheuerman, and Stacy~M Branham.
\newblock Gender recognition or gender reductionism? the social implications of embedded gender recognition systems.
\newblock In \emph{ACM Conference on Human Factors in Computing Systems}, pages 1--13, 2018.

\bibitem[H{\'e}bert-Johnson et~al.(2018)H{\'e}bert-Johnson, Kim, Reingold, and Rothblum]{hebert2018multicalibration}
Ursula H{\'e}bert-Johnson, Michael Kim, Omer Reingold, and Guy Rothblum.
\newblock Multicalibration: Calibration for the (computationally-identifiable) masses.
\newblock In \emph{International Conference on Machine Learning}, pages 1939--1948, 2018.

\bibitem[Hu et~al.(2021)Hu, Shen, Wallis, Allen-Zhu, Li, Wang, Wang, and Chen]{hu2021lora}
Edward~J Hu, Yelong Shen, Phillip Wallis, Zeyuan Allen-Zhu, Yuanzhi Li, Shean Wang, Lu Wang, and Weizhu Chen.
\newblock Lora: Low-rank adaptation of large language models.
\newblock 2021.

\bibitem[Karkkainen and Joo(2021)]{karkkainen2021fairface}
Kimmo Karkkainen and Jungseock Joo.
\newblock Fairface: Face attribute dataset for balanced race, gender, and age for bias measurement and mitigation.
\newblock In \emph{IEEE/CVF Winter Conference on Applications of Computer Vision}, pages 1548--1558, 2021.

\bibitem[Kim et~al.(2023)Kim, Kim, Shin, and Yoon]{kim2023stereotyping}
Eunji Kim, Siwon Kim, Chaehun Shin, and Sungroh Yoon.
\newblock De-stereotyping text-to-image models through prompt tuning.
\newblock \emph{Workshop on Challenges in Deployable Generative AI at Inter- national Conference on Machine Learning}, 2023.

\bibitem[Kim et~al.(2019)Kim, Ghorbani, and Zou]{kim2019multiaccuracy}
Michael~P Kim, Amirata Ghorbani, and James Zou.
\newblock Multiaccuracy: Black-box post-processing for fairness in classification.
\newblock In \emph{AAAI/ACM Conference on AI, Ethics, and Society}, pages 247--254, 2019.

\bibitem[Kim et~al.(2024)Kim, Na, Park, Jang, Kim, Kang, and Moon]{kim2024training}
Yeongmin Kim, Byeonghu Na, Minsang Park, JoonHo Jang, Dongjun Kim, Wanmo Kang, and Il-Chul Moon.
\newblock Training unbiased diffusion models from biased dataset.
\newblock \emph{International Conference on Learning Representations}, 2024.

\bibitem[King(2009)]{dlib09}
Davis~E. King.
\newblock Dlib-ml: A machine learning toolkit.
\newblock \emph{Journal of Machine Learning Research}, 10:\penalty0 1755--1758, 2009.

\bibitem[Kleiman(2022)]{kleiman2022proving}
Kathy Kleiman.
\newblock \emph{Proving ground: The untold story of the six women who programmed the world’s first modern computer}.
\newblock Hurst Publishers, 2022.

\bibitem[Li et~al.(2024{\natexlab{a}})Li, Kamko, Akhgari, Sabet, Xu, and Doshi]{li2024playground}
Daiqing Li, Aleks Kamko, Ehsan Akhgari, Ali Sabet, Linmiao Xu, and Suhail Doshi.
\newblock Playground v2. 5: Three insights towards enhancing aesthetic quality in text-to-image generation.
\newblock \emph{arXiv preprint arXiv:2402.17245}, 2024{\natexlab{a}}.

\bibitem[Li et~al.(2023)Li, Hu, Zhang, Zheng, Zhang, and Wang]{li2023fair}
Jia Li, Lijie Hu, Jingfeng Zhang, Tianhang Zheng, Hua Zhang, and Di Wang.
\newblock Fair text-to-image diffusion via fair mapping.
\newblock \emph{arXiv preprint arXiv:2311.17695}, 2023.

\bibitem[Li et~al.(2024{\natexlab{b}})Li, Liu, Zhang, Le, S{\"u}sstrunk, and Salzmann]{li2024controlling}
Shuangqi Li, Chen Liu, Tong Zhang, Hieu Le, Sabine S{\"u}sstrunk, and Mathieu Salzmann.
\newblock Controlling the fidelity and diversity of deep generative models via pseudo density.
\newblock \emph{Transactions on Machine Learning Research}, 2024{\natexlab{b}}.

\bibitem[Luccioni et~al.(2024)Luccioni, Akiki, Mitchell, and Jernite]{luccioni2024stable}
Sasha Luccioni, Christopher Akiki, Margaret Mitchell, and Yacine Jernite.
\newblock Stable bias: Evaluating societal representations in diffusion models.
\newblock \emph{Advances in Neural Information Processing Systems}, 36, 2024.

\bibitem[Luo et~al.(2024)Luo, Huang, Deng, Liu, Chen, and Liu]{luo2024bigbench}
Hanjun Luo, Haoyu Huang, Ziye Deng, Xuecheng Liu, Ruizhe Chen, and Zuozhu Liu.
\newblock Bigbench: A unified benchmark for social bias in text-to-image generative models based on multi-modal llm.
\newblock \emph{arXiv preprint arXiv:2407.15240}, 2024.

\bibitem[Luo et~al.(2023)Luo, Tan, Huang, Li, and Zhao]{luo2023latent}
Simian Luo, Yiqin Tan, Longbo Huang, Jian Li, and Hang Zhao.
\newblock Latent consistency models: Synthesizing high-resolution images with few-step inference.
\newblock \emph{arXiv preprint arXiv:2310.04378}, 2023.

\bibitem[Mack et~al.(2024)Mack, Qadri, Denton, Kane, and Bennett]{mack2024they}
Kelly~Avery Mack, Rida Qadri, Remi Denton, Shaun~K Kane, and Cynthia~L Bennett.
\newblock “they only care to show us the wheelchair”: disability representation in text-to-image ai models.
\newblock In \emph{CHI Conference on Human Factors in Computing Systems}, pages 1--23, 2024.

\bibitem[McDiarmid et~al.(1989)]{mcdiarmid1989method}
Colin McDiarmid et~al.
\newblock On the method of bounded differences.
\newblock \emph{Surveys in combinatorics}, 141\penalty0 (1):\penalty0 148--188, 1989.

\bibitem[Miao et~al.(2024)Miao, Wang, Wang, Yang, Wang, Qiu, and Liu]{miao2024training}
Zichen Miao, Jiang Wang, Ze Wang, Zhengyuan Yang, Lijuan Wang, Qiang Qiu, and Zicheng Liu.
\newblock Training diffusion models towards diverse image generation with reinforcement learning.
\newblock In \emph{IEEE/CVF Conference on Computer Vision and Pattern Recognition}, pages 10844--10853, 2024.

\bibitem[{Midjourney}(2024)]{midjourney}
{Midjourney}.
\newblock Midjourney.com, 2024.
\newblock Accessed Sep 7, 2024. \url{https://www.midjourney.com/home}.

\bibitem[Monteiro~Paes et~al.(2022)Monteiro~Paes, Long, Ustun, and Calmon]{monteiro2022epistemic}
Lucas Monteiro~Paes, Carol Long, Berk Ustun, and Flavio Calmon.
\newblock On the epistemic limits of personalized prediction.
\newblock \emph{Advances in Neural Information Processing Systems}, 35:\penalty0 1979--1991, 2022.

\bibitem[M{\"u}ller(1997)]{ipm1}
Alfred M{\"u}ller.
\newblock Integral probability metrics and their generating classes of functions.
\newblock \emph{Advances in applied probability}, 29\penalty0 (2):\penalty0 429--443, 1997.

\bibitem[Oesterling et~al.(2024)Oesterling, Mayrink~Verdun, Glynn, Long, Paes, Vithana, Cardone, and Calmon]{oesterling2024multi}
Alex Oesterling, Claudio Mayrink~Verdun, Alexander Glynn, Carol~Xuan Long, Lucas~Monteiro Paes, Sajani Vithana, Martina Cardone, and Flavio Calmon.
\newblock Multi-group proportional representation in retrieval.
\newblock In \emph{Advances in Neural Information Processing Systems}, 2024.

\bibitem[Orgad et~al.(2023)Orgad, Kawar, and Belinkov]{orgad2023editing}
Hadas Orgad, Bahjat Kawar, and Yonatan Belinkov.
\newblock Editing implicit assumptions in text-to-image diffusion models.
\newblock In \emph{IEEE/CVF International Conference on Computer Vision}, pages 7053--7061, 2023.

\bibitem[Ovalle et~al.(2023)Ovalle, Subramonian, Gautam, Gee, and Chang]{ovalle2023factoring}
Anaelia Ovalle, Arjun Subramonian, Vagrant Gautam, Gilbert Gee, and Kai-Wei Chang.
\newblock Factoring the matrix of domination: A critical review and reimagination of intersectionality in ai fairness.
\newblock In \emph{AAAI/ACM Conference on AI, Ethics, and Society}, pages 496--511, 2023.

\bibitem[Parihar et~al.(2024)Parihar, Bhat, Basu, Mallick, Kundu, and Babu]{parihar2024balancing}
Rishubh Parihar, Abhijnya Bhat, Abhipsa Basu, Saswat Mallick, Jogendra~Nath Kundu, and R~Venkatesh Babu.
\newblock Balancing act: Distribution-guided debiasing in diffusion models.
\newblock In \emph{IEEE/CVF Conference on Computer Vision and Pattern Recognition}, pages 6668--6678, 2024.

\bibitem[Pernias et~al.(2023)Pernias, Rampas, Richter, Pal, and Aubreville]{pernias2023wurstchen}
Pablo Pernias, Dominic Rampas, Mats~L Richter, Christopher~J Pal, and Marc Aubreville.
\newblock W{\"u}rstchen: An efficient architecture for large-scale text-to-image diffusion models.
\newblock \emph{arXiv preprint arXiv:2306.00637}, 2023.

\bibitem[Peyr{\'e} et~al.(2019)Peyr{\'e}, Cuturi, et~al.]{peyre2019computational}
Gabriel Peyr{\'e}, Marco Cuturi, et~al.
\newblock Computational optimal transport: With applications to data science.
\newblock \emph{Foundations and Trends{\textregistered} in Machine Learning}, 11\penalty0 (5-6):\penalty0 355--607, 2019.

\bibitem[Podell et~al.(2024)Podell, English, Lacey, Blattmann, Dockhorn, M{\"u}ller, Penna, and Rombach]{podellsdxl}
Dustin Podell, Zion English, Kyle Lacey, Andreas Blattmann, Tim Dockhorn, Jonas M{\"u}ller, Joe Penna, and Robin Rombach.
\newblock Sdxl: Improving latent diffusion models for high-resolution image synthesis.
\newblock In \emph{International Conference on Learning Representations}, 2024.

\bibitem[Raghavan()]{geminiarticle2}
Prabhakar Raghavan.
\newblock Gemini image generation got it wrong. we'll do better.
\newblock \url{https://blog.google/products/gemini/gemini-image-generation-issue/}.
\newblock Accessed: 2024-11-09.

\bibitem[Ramesh et~al.(2021)Ramesh, Pavlov, Goh, Gray, Voss, Radford, Chen, and Sutskever]{ramesh2021zero}
Aditya Ramesh, Mikhail Pavlov, Gabriel Goh, Scott Gray, Chelsea Voss, Alec Radford, Mark Chen, and Ilya Sutskever.
\newblock Zero-shot text-to-image generation.
\newblock In \emph{International conference on machine learning}, pages 8821--8831. {PMLR}, 2021.

\bibitem[Robertson()]{geminiarticle}
Adi Robertson.
\newblock Google apologizes for ‘missing the mark’ after gemini generated racially diverse nazis.
\newblock \url{https://www.theverge.com/2024/2/21/24079371/google-ai-gemini-generative-inaccurate-historical}.
\newblock Accessed: 2024-11-09.

\bibitem[Rombach et~al.(2022)Rombach, Blattmann, Lorenz, Esser, and Ommer]{rombach2022high}
Robin Rombach, Andreas Blattmann, Dominik Lorenz, Patrick Esser, and Bj{\"o}rn Ommer.
\newblock High-resolution image synthesis with latent diffusion models.
\newblock In \emph{IEEE/CVF Conference on Computer Vision and Pattern Recognition}, pages 10684--10695, 2022.

\bibitem[Shen et~al.(2023)Shen, Du, Pang, Lin, Wong, and Kankanhalli]{shen2023finetuning}
Xudong Shen, Chao Du, Tianyu Pang, Min Lin, Yongkang Wong, and Mohan Kankanhalli.
\newblock Finetuning text-to-image diffusion models for fairness.
\newblock \emph{arXiv preprint arXiv:2311.07604}, 2023.

\bibitem[Srinivasan et~al.(2024)Srinivasan, Schumann, Sinha, Madras, Olanubi, Beutel, Ricco, and Chen]{srinivasan2024generalized}
Hansa Srinivasan, Candice Schumann, Aradhana Sinha, David Madras, Gbolahan~Oluwafemi Olanubi, Alex Beutel, Susanna Ricco, and Jilin Chen.
\newblock Generalized people diversity: Learning a human perception-aligned diversity representation for people images.
\newblock In \emph{ACM Conference on Fairness, Accountability, and Transparency}, pages 797--821, 2024.

\bibitem[Sriperumbudur et~al.(2012)Sriperumbudur, Fukumizu, Gretton, Sch{\"o}lkopf, and Lanckriet]{ipm2}
Bharath~K Sriperumbudur, Kenji Fukumizu, Arthur Gretton, Bernhard Sch{\"o}lkopf, and Gert~RG Lanckriet.
\newblock On the empirical estimation of integral probability metrics.
\newblock \emph{Electronic Journal of Statistics}, 6:\penalty0 1550--1599, 2012.

\bibitem[{Stabel Diffusion 2.1}(2022)]{SD2.1}
{Stabel Diffusion 2.1}.
\newblock Stable diffusion v2.1 and dreamstudio updates 7-dec 22, 2022.
\newblock Accessed September 10, 2024. \url{https://stability.ai/news/stablediffusion2-1-release7-dec-2022}.

\bibitem[Wan et~al.(2024)Wan, Subramonian, Ovalle, Lin, Suvarna, Chance, Bansal, Pattichis, and Chang]{wan2024survey}
Yixin Wan, Arjun Subramonian, Anaelia Ovalle, Zongyu Lin, Ashima Suvarna, Christina Chance, Hritik Bansal, Rebecca Pattichis, and Kai-Wei Chang.
\newblock Survey of bias in text-to-image generation: Definition, evaluation, and mitigation.
\newblock \emph{arXiv preprint arXiv:2404.01030}, 2024.

\bibitem[Wang et~al.(2024)Wang, Gui, Negrea, and Veitch]{wang2024concept}
Zihao Wang, Lin Gui, Jeffrey Negrea, and Victor Veitch.
\newblock Concept algebra for (score-based) text-controlled generative models.
\newblock \emph{Advances in Neural Information Processing Systems}, 36, 2024.

\bibitem[Yesiltepe et~al.(2024)Yesiltepe, Akdemir, and Yanardag]{yesiltepe2024mist}
Hidir Yesiltepe, Kiymet Akdemir, and Pinar Yanardag.
\newblock Mist: Mitigating intersectional bias with disentangled cross-attention editing in text-to-image diffusion models.
\newblock \emph{arXiv preprint arXiv:2403.19738}, 2024.

\bibitem[Zhang et~al.(2023)Zhang, Chen, Chai, Wu, Lagun, Beeler, and De~la Torre]{zhang2023iti}
Cheng Zhang, Xuanbai Chen, Siqi Chai, Chen~Henry Wu, Dmitry Lagun, Thabo Beeler, and Fernando De~la Torre.
\newblock Iti-gen: Inclusive text-to-image generation.
\newblock In \emph{IEEE/CVF International Conference on Computer Vision}, pages 3969--3980, 2023.

\bibitem[Zhao et~al.(2022)Zhao, Huang, Liu, Yu, Liu, Russakovsky, and Anandkumar]{zhao2022scaling}
Eric Zhao, De-An Huang, Hao Liu, Zhiding Yu, Anqi Liu, Olga Russakovsky, and Anima Anandkumar.
\newblock Scaling fair learning to hundreds of intersectional groups.
\newblock 2022.

\end{thebibliography}
